\documentclass[10pt,twocolumn,letterpaper]{article}

\usepackage{cvpr}              
\usepackage[accsupp]{axessibility}  
\usepackage{graphicx}
\usepackage{amsmath}
\usepackage{amssymb}
\usepackage{booktabs}
\usepackage{bbm}

\usepackage[ruled,vlined]{algorithm2e}

\newcommand{\comment}[1]{}

\usepackage[pagebackref,breaklinks,colorlinks]{hyperref}

\usepackage[pagebackref,breaklinks,colorlinks]{hyperref}

\DeclareMathOperator{\shuffle}{Shuffle}
\DeclareMathOperator{\concat}{Concat}

\DeclareMathOperator{\xent}{H}

\usepackage[capitalize]{cleveref}
\crefname{section}{Sec.}{Secs.}
\Crefname{section}{Section}{Sections}
\Crefname{table}{Table}{Tables}
\crefname{table}{Tab.}{Tabs.}

\def\cvprPaperID{7363} 
\def\confName{CVPR}
\def\confYear{2022}

\begin{document}

\title{
\textsc{UniCon}: Combating Label Noise Through Uniform Selection and Contrastive Learning\\

}


\author{Nazmul Karim\IEEEauthorrefmark2 \hspace{2mm} Mamshad Nayeem Rizve\IEEEauthorrefmark3 \hspace{2mm} Nazanin Rahnavard\IEEEauthorrefmark2 \hspace{2mm} Ajmal Mian\IEEEauthorrefmark4 \hspace{2mm} Mubarak Shah\IEEEauthorrefmark3 \and
{\normalsize \IEEEauthorrefmark2Department of Electrical and Computer Engineering, UCF, USA}\and{\normalsize \IEEEauthorrefmark3Center for Research in Computer Vision, UCF, USA}\and{\normalsize \IEEEauthorrefmark4Department of Computer Science and Software Engineering, UWA, Australia}\and
\texttt{\small \{nazmul.karim18, nayeemrizve\}@knights.ucf.edu, nazanin.rahnavard@ucf.edu}\and \texttt{\small ajmal.mian@uwa.edu.au,
shah@crcv.ucf.edu}}

\maketitle

\begin{abstract}
\vspace{-4mm}

Supervised deep learning methods require a large repository of annotated data; hence, label noise is inevitable.
Training with such noisy data negatively impacts the generalization performance of deep neural networks. To combat
label noise, recent state-of-the-art methods employ some
sort of sample selection mechanism to select a possibly
clean subset of data. Next, an off-the-shelf semi-supervised
learning method is used for training where rejected samples
are treated as unlabeled data. Our comprehensive analysis
shows that current selection methods disproportionately select samples from easy (fast learnable) classes while rejecting
those from relatively harder ones. This creates class imbalance in the selected clean set and in turn, deteriorates performance under high label noise. In this work, we propose
UNICON, a simple yet effective sample selection method
which is robust to high label noise. To address the disproportionate selection of easy and hard samples, we introduce a Jensen-Shannon divergence based uniform selection
mechanism which does not require any probabilistic modeling and hyperparameter tuning. We complement our selection method with contrastive learning to further combat
the memorization of noisy labels. Extensive experimentation on
multiple benchmark datasets demonstrates the effectiveness
of UNICON; we obtain an 11.4\% improvement over the current state-of-the-art on CIFAR100 dataset with a 90\% noise
rate. Our code is publicly available.\footnote{\url{https://github.com/nazmul-karim170/UNICON-Noisy-Label}}

\end{abstract}

\vspace{-2mm}
\section{Introduction}
\label{sec:intro}
\vspace{-2mm}






Deep neural networks (DNNs) have proven to be highly effective in solving various computer vision tasks \cite{krizhevsky2012imagenet,noh2015learning, diaz2009integration,redmon2016look, yao2021non, joseph2021towards,wang2021contrastive, song2021addersr, srinivas2021bottleneck}. Most state-of-the-art (SOTA) methods require supervised training with a large pool of annotated data~\cite{deng2009imagenet,xiao2015learning, li2017webvision, lin2014microsoft, bossard2014food}. Collecting and manually annotating such data is challenging and oftentimes very expensive. Most large-scale data collection techniques rely on open-source web data that can be automatically annotated using search engine queries and user tags \cite{makadia2008new,tsai2008automatically}. This annotation scheme inevitably introduces label noise \cite{xiao2015learning, li2017webvision}. Training with such noisy labels is challenging since DNNs can effectively memorize arbitrary (noisy) labels over the course of training \cite{arpit2017closer}. Combating label noise is one of the fundamental problems in deep learning \cite{xie2020self,wang2019co,wang2019co, hu2019noise,shen2020noise,li2020learning, xue2019robust, yang2021joint, yu2021divergence,oh2021background}, and is the focus of this study.

\begin{figure}[t]
\centering
\includegraphics[width=0.9\linewidth]{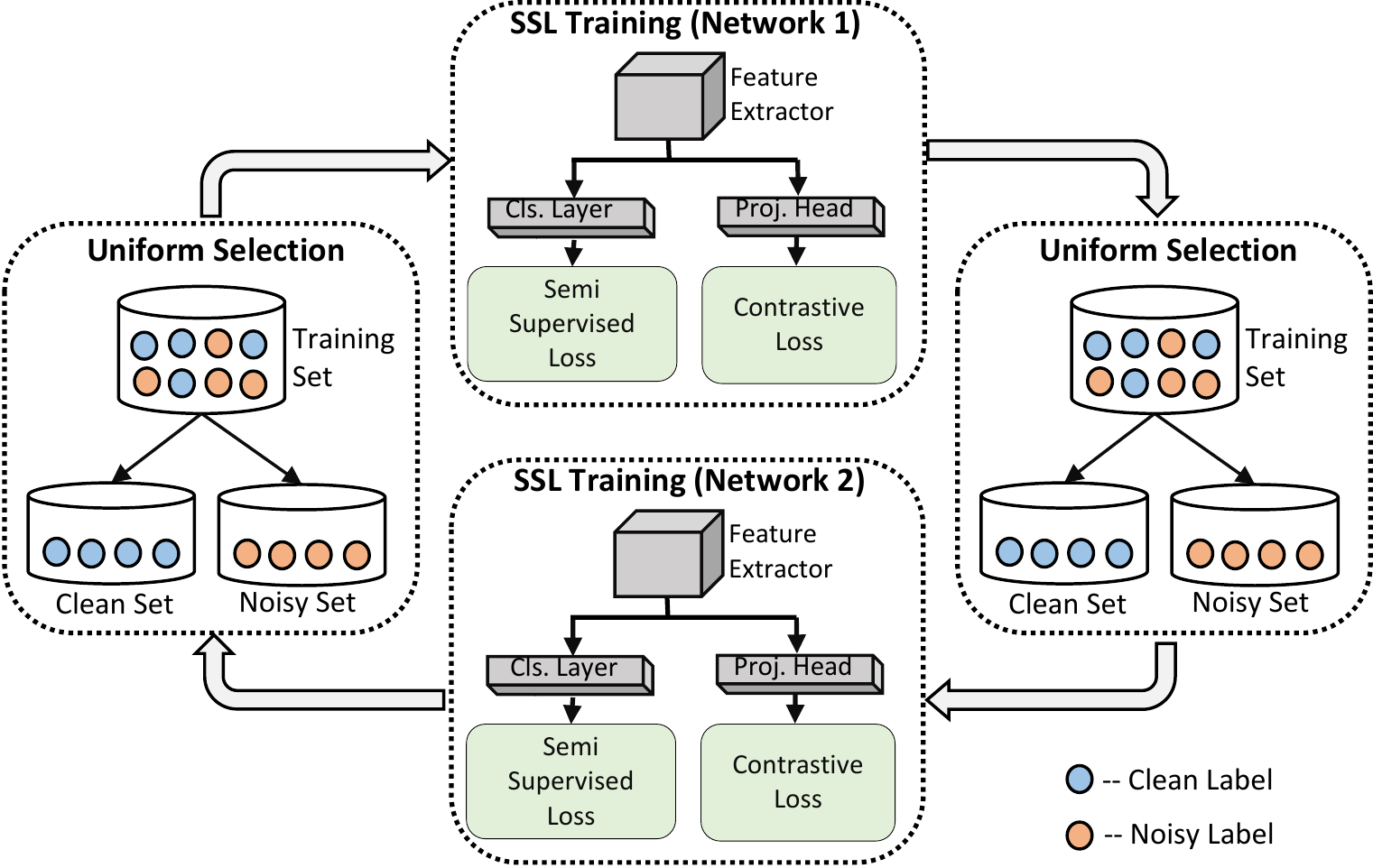}
\vspace{-2mm}
\caption{\textit{\textsc{UniCon} training overview:} At each iteration, we employ a uniform selection technique to partition the training set into clean and noisy sets. Upon separation, we perform SSL-training with an additional contrastive loss function. The uniform selection and subsequent SSL-training of two networks (with same architecture) is repeated until convergence.}

\label{fig:framework}
\vspace{-2mm}
\end{figure}
\begin{table}[t]
\centering
\small
\scalebox{1}{
\begin{tabular}{l|cccc}
\toprule
Noise Rate (\%) & 90\% & 92\% & 95\% & 98\%  \\ \midrule
DMix \cite{li2020dividemix}  & 76.08 & 57.62  &  51.28  & 17.18\\ 
\textsc{UniCon} (Ours) & \textbf{90.81} & \textbf{87.61} & \textbf{80.82} & \textbf{50.63}\\
 \bottomrule
\end{tabular}
}
\vspace{-1mm}
\caption{Classification performance (\%) of the proposed method on CIFAR10 under severe label noise.
}
\label{tab:severe}
\vspace{-4mm}
\end{table}

Training with noisy label data has been the subject of many recent studies ~\cite{han2018co,jiang2020beyond,liu2020peer,qu2021dat,zhang2021learning,sharma2020noiserank}. Existing techniques can be categorized into two dominant groups: i) label correction, \cite{goldberger2016training,patrini2017making} and ii) sample separation~\cite{han2018co,yu2019does,li2020dividemix}. 
The former approach requires the estimation of noise transition matrix, which is hard to estimate for high number of classes and in high noise scenarios. 
The latter approach tries to filter out the noisy samples from the clean ones based on the small-loss criterion~\cite{li2020dividemix}, where the low-loss samples are assumed to have clean labels. Next, an off-the-shelf semi-supervised learning (SSL) technique \cite{berthelot2019mixmatch, sohn2020fixmatch, rizve2021in, tarvainen2017mean} is used for training where the selected noisy samples are treated as unlabeled data.  However, the selection process is usually biased towards easy classes as clean samples from the hard classes (e.g. cats and dogs can be considered as hard classes in CIFAR10 \cite{krizhevsky2009learning}) may produce high-loss values. This is more prominent at the early stage of training and can introduce class-disparity among the selected clean samples. Severe class-imbalance may lead to poor precision of sample selection, hence, sub-par classification performance.  

In this work, we revamp the selection process from a more fundamental perspective. Our goal is to simplify the selection process by introducing an effective and scalable Jensen-Shannon divergence based sample separation mechanism. To address the disproportionate selection of easy and hard samples, we enforce a class-balance prior by selecting an equal number of clean samples from each class. Such a prior improves the overall quality of pseudo-labels, and hence, significantly boosts the performance of subsequent semi supervised learning-based training. In addition, we opt to employ unsupervised contrastive learning (CL) because of its inherent resistance (as labels are not required for training) to label noise memorization. We empirically show that unsupervised feature learning lowers memorization risk and improves the sample separation performance; especially under severe noise levels. We call this combined technique of \textsc{Uni}form selection and \textsc{Con}trastive learning \textsc{UniCon} (shown in Fig. \ref{fig:framework}), which is found to be effective even in the presence of very high label noise (see Table \ref{tab:severe}). Our contributions are summarized as follows:



\begin{itemize}
    \vspace{-2mm}
    \item We propose a simple yet effective uniform selection mechanism that ensures class-balancing among the selected clean samples. Through empirical analysis, we observe that class-uniformity helps in generating higher quality pseudo-labels for samples from all classes irrespective of their difficulty level. 
    
    \vspace{-2mm}
    \item We further minimize the risk of label noise memorization by performing unsupervised feature learning using contrastive loss. This in turn boosts the sample separation performance.
    
    \vspace{-2mm}
    \item Our extensive experimentation demonstrates that \textsc{UniCon} achieves significant performance improvement over state-of-the-art methods, especially on datasets with severe label noise.  
\end{itemize}


\begin{figure*}[htb]
  \centering
  \begin{subfigure}{0.245\linewidth}
    \includegraphics[width=1\linewidth]{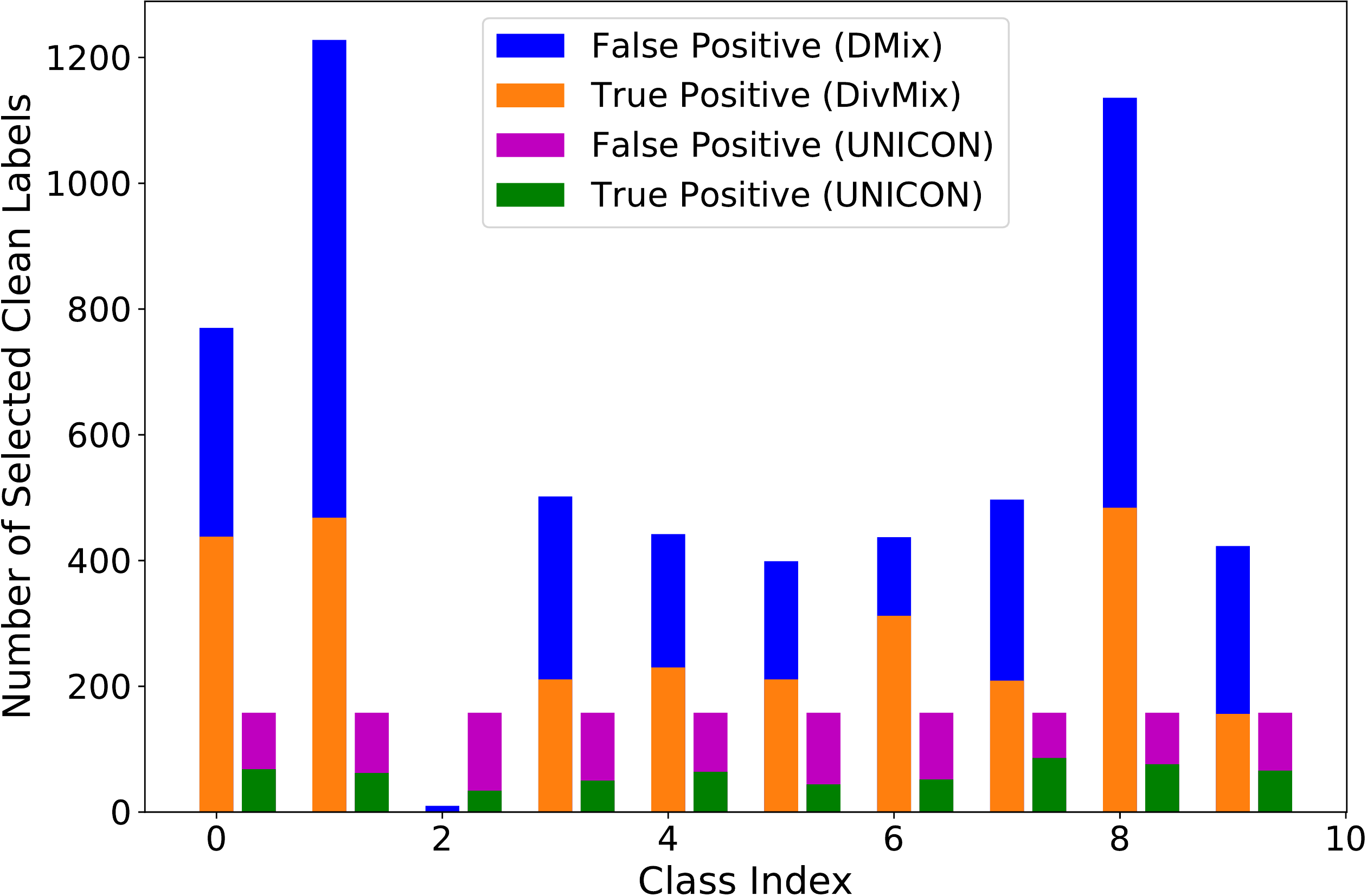}
    \caption{ \scriptsize Selected clean set after warmup.}
    \label{fig:main-sel}
  \end{subfigure}
    \begin{subfigure}{0.245\linewidth}
    \includegraphics[width=0.9\linewidth]{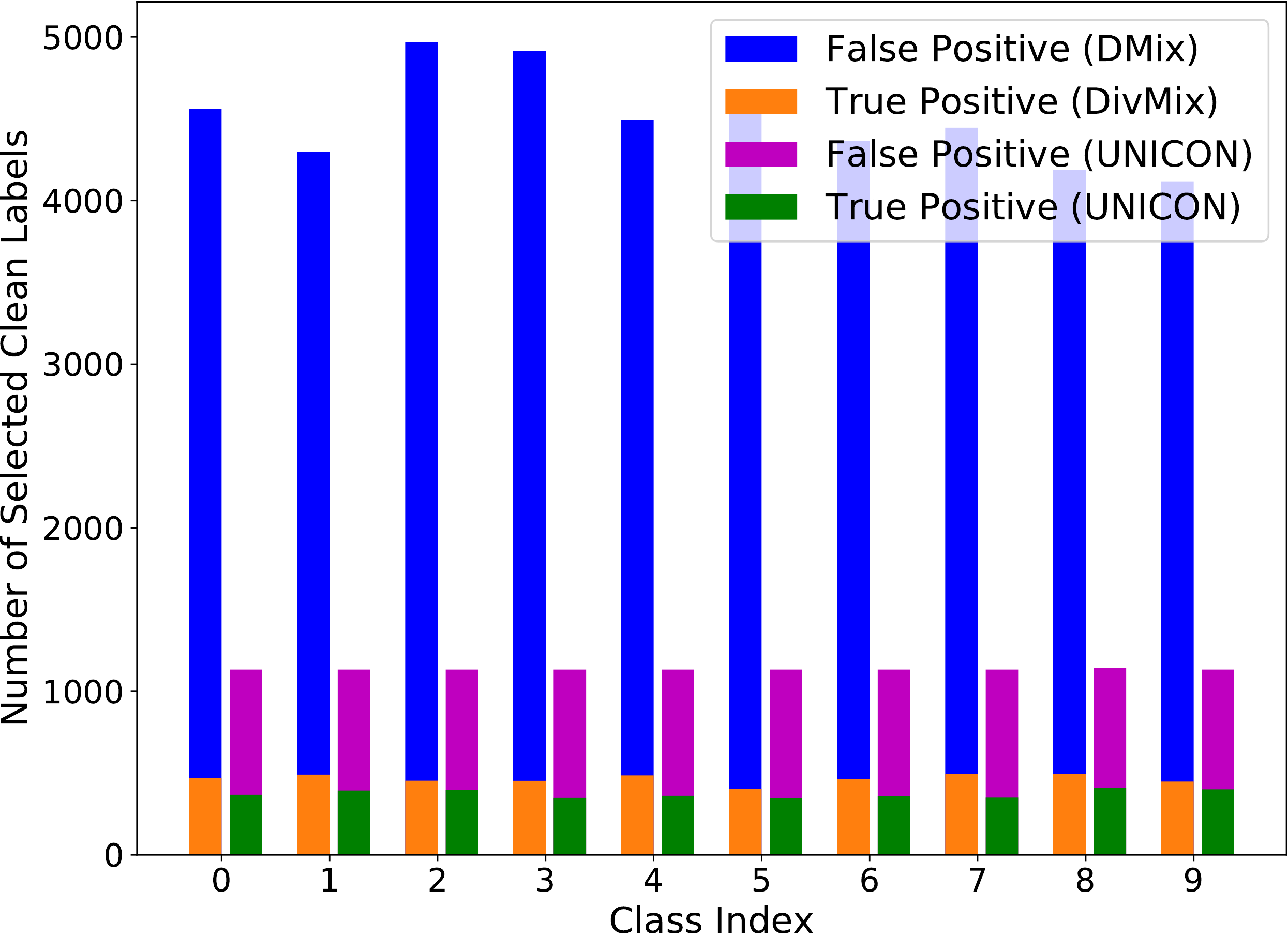}
    \caption{ \scriptsize Selected clean set after 100 epochs.}
    \label{fig:main-sel_100}
  \end{subfigure}
  \hfill
  \begin{subfigure}{0.245\linewidth}
    \includegraphics[width=0.97\linewidth]{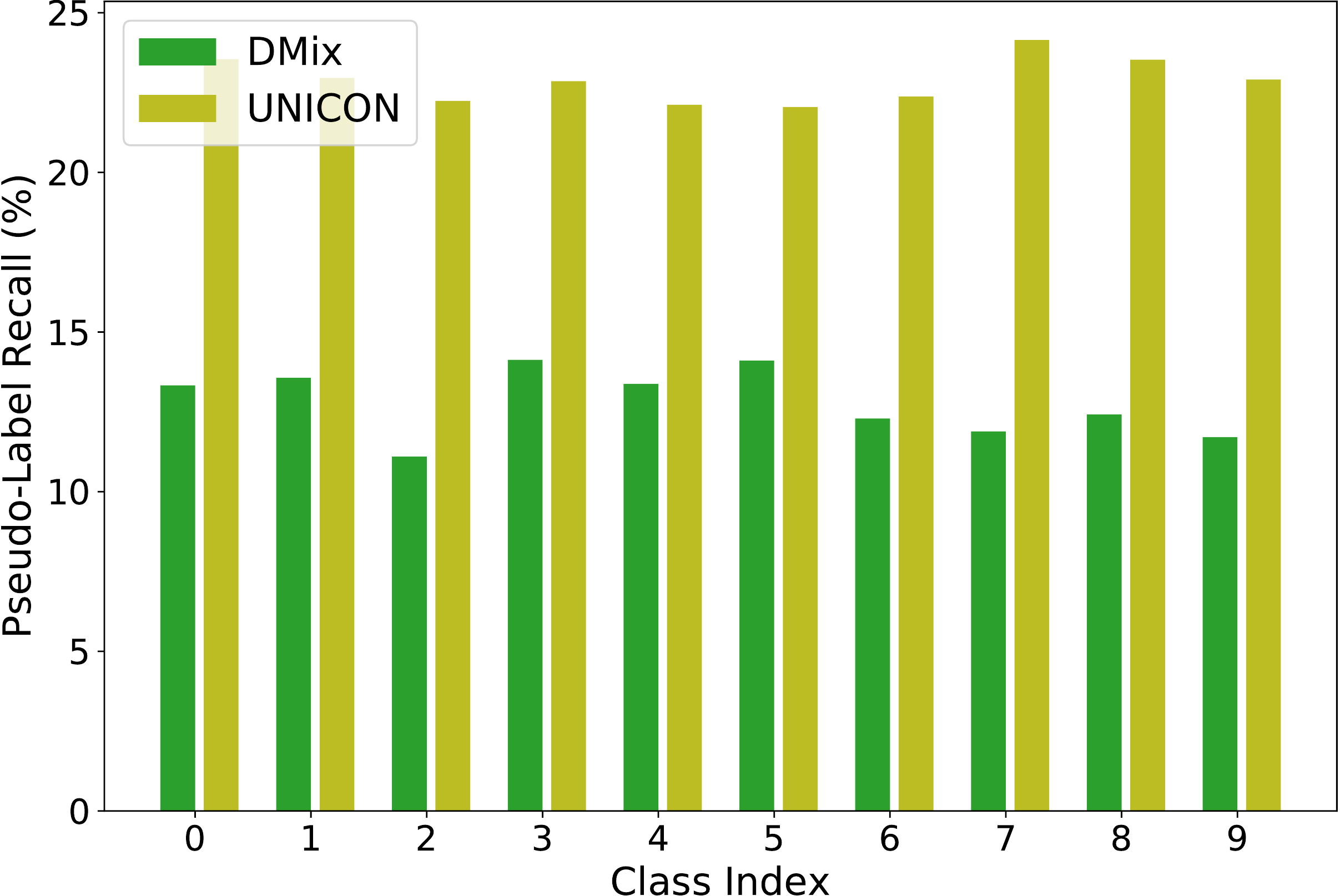}
    \caption{ \scriptsize Pseudo-label recall after 100 epochs.}
    \label{fig:main-recall}
  \end{subfigure}
    \begin{subfigure}{0.245\linewidth}
    \includegraphics[width=0.93\linewidth]{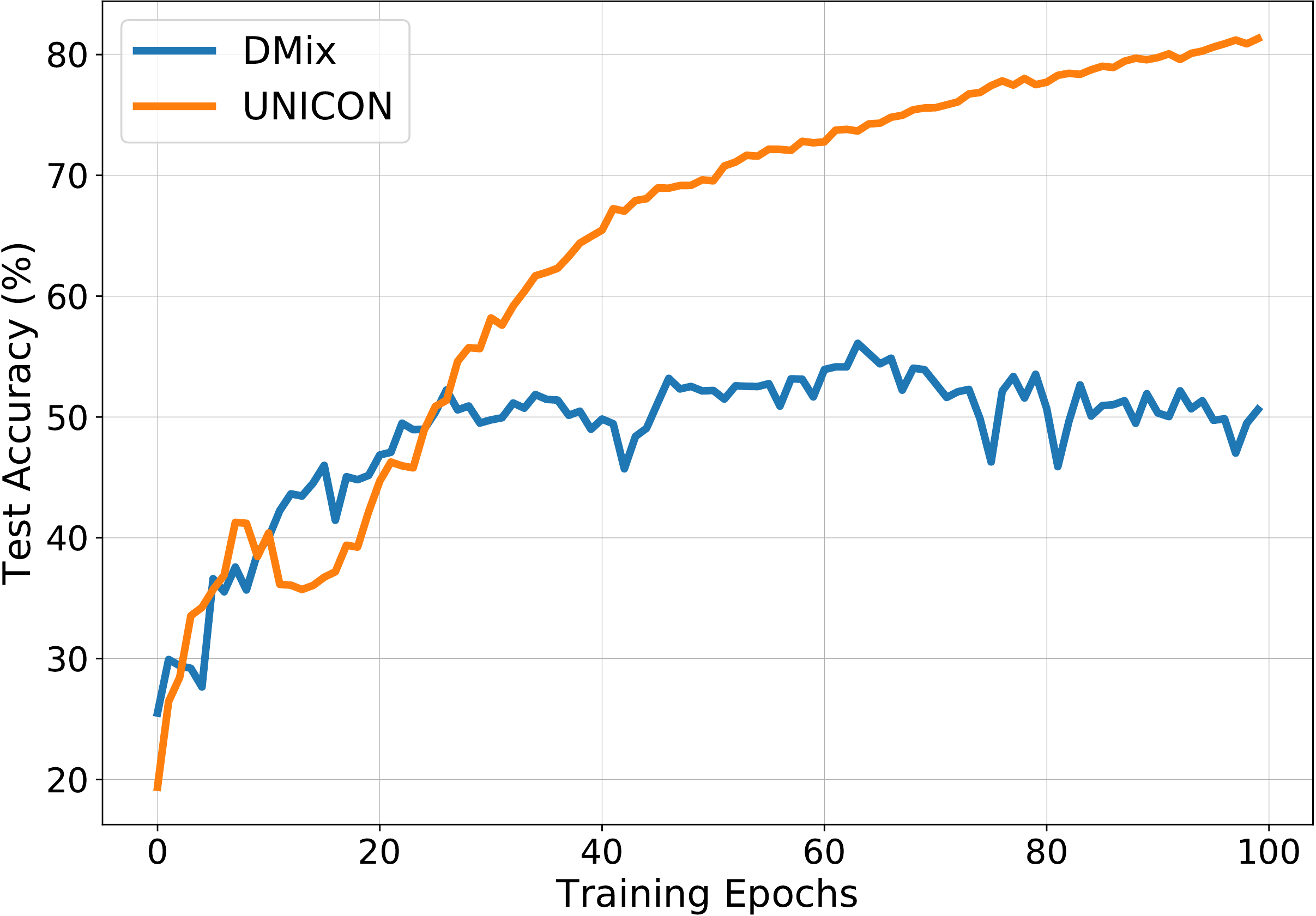}
    \caption{\scriptsize Test accuracy (\%).}
    \label{fig:main-acc}
  \end{subfigure}
  \vspace{-1mm}
  \caption{A case of uniform and non-uniform selection for CIFAR10 under 90\% noise rate. (a) Class distribution in $\mathbb{D}_{clean}$ after warmup (10 epochs of training). For each class index, the left and right bars indicate non-uniform (DMix~\cite{li2020dividemix}) and uniform selection (\textsc{UniCon}), respectively. (b) Class distribution after 100 epochs. \textsc{UniCon} selects clean samples with higher precision. (c) pseudo-label recall (\%) after 100 epochs of training. Uniform selection criteria along with contrastive feature learning helps generating higher quality pseudo-labels with better recall. (d) This in turn boosts the test accuracy significantly.}
  \label{fig:main}
  \vspace{-2mm}
\end{figure*}
\section{Related Work} \label{sec:related work}
Noisy label training has been studied extensively in recent works~\cite{liu2021noiseresistant, zheng2021meta, nishi2021augmentation,wang2018iterative,li2019learning}. Wei et al.~\cite{wei2020combating} proposed a regularization technique to learn from noisy labels. Another method called MentorNet \cite{jiang2018mentornet} trains a student network by generating pseudo-labels using a pre-trained/mentor network. 
Based on their relationship in the feature space, Meta-cleaner \cite{zhang2019metacleaner} learns the confidence scores of noisy samples which are then used for obtaining cleaner representations. To deal with noisy labels, \cite{ma2018dimensionality, tanaka2018joint, yi2019probabilistic} gradually adjust the data labels based on the predicted labels given by the network. 
Some noisy label methods are based on loss correction \cite{goldberger2016training,hendrycks2018using, patrini2017making} and noise-tolerant loss functions \cite{chang2018active, zhang2018generalized}. In \cite{hendrycks2018using}, a noise transition matrix was estimated by correcting the loss obtained by a DNN trained on a noisy dataset. However, the performance of these methods deteriorates under high noise rates and large number of classes.
Other approaches rely on the separation of clean samples from the noisy samples \cite{ding2018semi, northcutt2017learning, han2018co, li2020dividemix, yu2019does, tanaka2018joint}. A notable difference between these methods is the selection criteria of clean samples. A selection technique was proposed in~\cite{ding2018semi} that utilizes prediction likelihoods to obtain separation. 

Co-teaching \cite{han2018co} opts to train two networks simultaneously such that one network separates clean samples for the other network based on the small-loss criterion. The small-loss criterion suggests that samples with smaller loss tend to have clean labels. Therefore, one could separate samples on the training set based on their loss-values. 
DMix \cite{li2020dividemix} proposed a hybrid framework to separate samples and uses a SSL technique~\cite{zhang2018mixup} to concurrently train two networks. A modified training scheme for \cite{li2020dividemix} was proposed in \cite{nishi2021augmentation}. However, even for the same dataset, these methods employ different training settings and constraints under different noise rates and types. This limits their practical applications as prior knowledge of noise rate may not be available. Recently, a joint semi-supervised and contrastive learning-based technique was proposed in MOIT \cite{ortego2021multiobjective}. Jo-SRC \cite{yao2021jo} initially partitions the samples into clean and noisy sets before detecting in-distribution (ID) and OOD samples in the noisy set. 
However, it requires manual threshold adjustment for the separation during different epochs of the training. Furthermore, both \cite{yao2021jo} and \cite{ortego2021multiobjective} struggle to achieve good performance under high noise rates.

In contrast, our proposed method can handle severe label noise and requires minimal to no change in the hyper-parameter settings under different label-noise scenarios (e.g. different noise rates, noise types etc.). We show how a minimalistic approach to the selection process can boost the classification performance significantly beating the sate-of-the-art methods in most cases. Furthermore, we achieve comparable performance to SOTA across different datasets which hints at the generalizability of our method. 

\section{Background} \label{sec:Method}
Let $ \mathbb{D} = \{\mathcal{X},\mathcal{Y}\}= {\{(\mathbf{x}_0, \mathbf{y}_0),(\mathbf{x}_1, \mathbf{y}_1), \ldots,
(\mathbf{x}_N , \mathbf{y}_N )\}}$ denote the training set, where $\mathbf{x}_i$
is an image and $\mathbf{y}_i$ 
is the corresponding ground-truth label,  
and $N$ is the total number of training samples. We instantiate the DNN model with a feature extractor (CNN backbone), $\mathbf{f}(.;\theta)$, with parameters $\theta$; a classification layer, $\mathbf{h}(.;\phi)$, with parameters $\phi$, and a projection head, $\mathbf{g}(.,\psi)$, with parameters $\psi$ for incorporating contrastive learning.
For supervised training with ground-truth labels, we minimize cross-entropy (CE) loss, $\mathcal{L}_{CE}$, over the entire training set $\mathbb{D}$,
\begin{equation}
    \vspace{-0.01cm}
    \mathcal{L}_{CE} = - \frac{1}{N}\sum_{i=1}^{N}{\mathbf{y}_i^T \log\mathbf{\Hat{y}}_i}, 
    \vspace{-1mm}
\end{equation}

\noindent where $\mathbf{\Hat{y}}_i= \mathrm{softmax}(\mathbf{h}(\mathbf{f}(\mathbf{x}_i;\theta);\phi))$ is the softmax probability score of the network prediction corresponding to $\mathbf{x}_i$.

In this work, we consider the training set to be noisy i.e.~some images are incorrectly labeled. 
It has been demonstrated that DNNs learn simpler patterns before memorizing the noisy labels \cite{arpit2017closer}. Several studies \cite{li2020dividemix, han2018co} utilize this observation and try to separate the clean samples from the noisy ones at the early stage of training. Such a separation scheme partitions the dataset into a clean subset, $\mathbb{D}_{clean}$, and a noisy subset, $\mathbb{D}_{noisy} = \mathbb{D} \setminus \mathbb{D}_{clean}$.
After that, $\mathbb{D}_{clean}$ can be used for standard supervised training. To mitigate the impact of label noise, samples from $\mathbb{D}_{noisy}$ can be used for training without the corresponding noisy ground-truth labels. This training is generally performed in a semi-supervised manner where pseudo-labels are generated for the samples in $\mathbb{D}_{noisy}$.

We conduct extensive empirical analysis to investigate the effectiveness of partitioning the dataset into $\mathbb{D}_{clean}$, and $\mathbb{D}_{noisy}$ subsets. We find that the typical construction of $\mathbb{D}_{clean}$ creates disparity or \emph{imbalance} among classes~\cite{li2020dividemix,han2018co,yao2021jo}. Fig.~\ref{fig:main-sel} (left bars) depicts such a case where the $\mathbb{D}_{clean}$ for noisy CIFAR10 (90\% noise rate) contain class imbalance when we employ a recently proposed method, DMix~\cite{li2020dividemix}. To be specific, we observe that $1228$ samples are selected from class-$1$, whereas only $10$ samples from class-$2$ are selected. However, the imbalance among true positives (TPs) are of particular importance 
as the quality of pseudo-labels for $\mathbb{D}_{noisy}$ relies heavily on them. Methods such as  \cite{li2020dividemix} attempt to address this issue by selecting more clean samples which in turn increases the false positive or noisy labels count (Fig.~\ref{fig:main-sel_100} (left bars)) while drastically decreasing the precision. As the selected clean set $\mathbb{D}_{clean}$ contains many false positives, supervised training on such a set leads to memorization. Consequently, the recall of the subsequent pseudo-labels 
drops drastically; as shown by the pseudo-label recall in Fig.~\ref{fig:main-recall} (left bars).
In this way, the selection mechanism negatively impacts the SSL-Training and reduces the average classification accuracy (Fig.~\ref{fig:main-acc}).

\begin{figure*}[htb]
\centering
\includegraphics[width=0.9\linewidth]{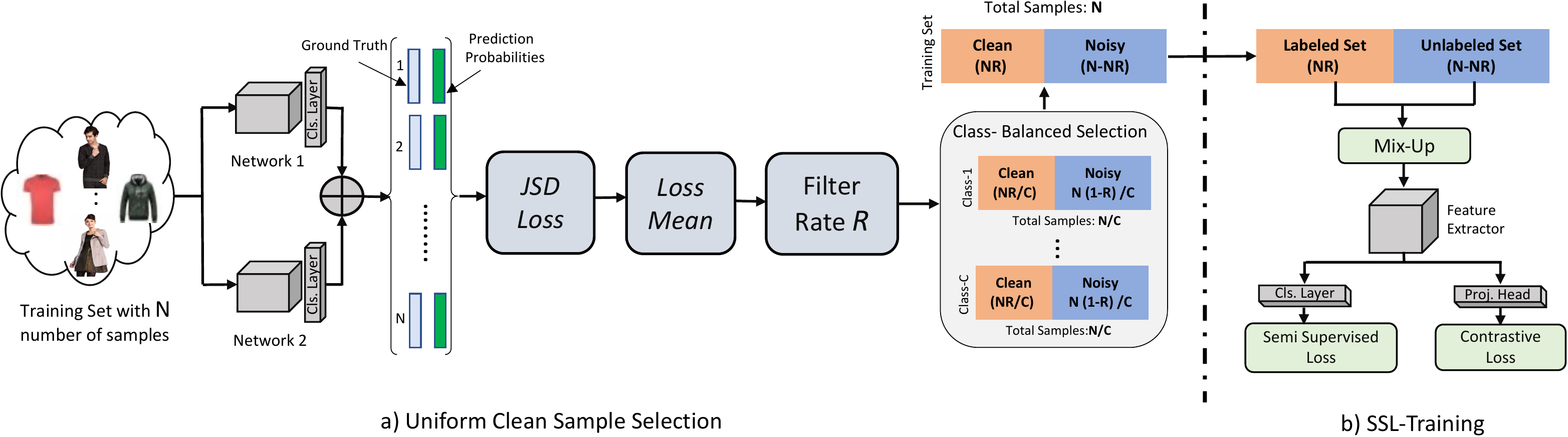}
\vspace{-1mm}
\caption{{\it Proposed Selection Mechanism and SSL-Training:} a) For selection, we ensemble the predictions of both networks to calculate JSD. After estimating the filter rate $R$ from JSD distribution, we take equal number ($NR/C$) of samples from each class. b) We consider separated clean and noisy sets as labeled and unlabeled data only to employ Mix-up \cite{zhang2017mixup} based SSL-training with contrastive loss. On top of classification (Cls.) layer, we add a projection (Proj.) head to facilitate contrastive learning. We train both networks sequentially.}

\label{fig:sep}
\vspace{-1mm}
\end{figure*}



We propose to address these problems by a simple and effective technique of uniform selection (Fig.~\ref{fig:main-sel} (right bars)). Furthermore, we employ contrastive feature learning to learn better unsupervised features irrespective of the quality of ground-truth or pseudo-labels. 
Details of our proposed method are presented in the following section.  
\vspace{-1mm}
\section{Proposed Method}
We propose \textsc{UniCon} with a unique sample-selection approach as well as simple but effective modification to the SSL-Training.
\textsc{UniCon} improves precision (Fig.~\ref{fig:main-sel_100} (right bars)) as well as  pseudo-label recall (Fig.~\ref{fig:main-recall} (right bars)) over training. Fig.~\ref{fig:main-acc} shows that our hybrid framework of uniform selection and SSL training improves the classification performance significantly. Next, we present our uniform sample selection strategy in Sec.~\ref{sec:uni_sel}, and our proposed SSL training method with contrastive learning in Sec~\ref{sec:ssl_train}.


\subsection{Uniform Sample Selection}
\label{sec:uni_sel}
During the partitioning of $\mathbb{D}$, we opt to enforce class-balancing in $\mathbb{D}_{clean}$ by selecting/filtering $R$ portion of samples from {\it each class}, where we define $R$ as the filter rate. Fig.~\ref{fig:sep} shows our proposed selection mechanism in which we feed $\mathbb{D}$ to two networks with parameters $(\theta^{(1)}, \phi^{(1)}, \psi^{(1)})$ and $(\theta^{(2)}, \phi^{(2)}, \psi^{(2)})$. 
For $\mathbf{x}_i$, the average 
prediction probabilities from both networks can be denoted as $\mathbf{p}_i = [\mathbf{p}_{i}^{1}, \mathbf{p}_{i}^{2}, \ldots,\mathbf{p}_{i}^{C}]$, and the corresponding ground-truth label as $\mathbf{y}_i = [\mathbf{y}_{i}^{1}, \mathbf{y}_{i}^{2}, \ldots,\mathbf{y}_{i}^{C}]$; here, $C$ is the total number of classes.
To construct the clean, $\mathbb{D}_{clean}$, and noisy, $\mathbb{D}_{noisy}$, subsets, we compute the disagreement/divergence between the ground-truth labels, $\mathbf{y}_i$, and the predicted probabilities, $\mathbf{p}_i$. To this end, we use Jensen-Shannon divergence (JSD), $d_i$, as a measure of disagreement. The JSD is defined as, 
\begin{equation}\label{eq:JSD}
\begin{split}
    d_i & = \mathrm{JSD}(\mathbf{y}_i,\mathbf{p}_i) \\
       & = \frac{1}{2} \mathrm{KLD}(\mathbf{y}_i||\frac{\mathbf{y}_i+\mathbf{p}_i}{2}) + \frac{1}{2} \mathrm{KLD}(\mathbf{p}_i||\frac{\mathbf{y_i}+\mathbf{p}_i}{2}),
\end{split}
\end{equation}
where $\mathrm{KLD}(.)$ is the Kullback-Leibler divergence function.

Previous works use different divergence measures to construct the clean and noisy subsets. Authors in \cite{han2018co, li2020dividemix} apply CE loss-based divergence measure for selection. \cite{li2020dividemix} uses a similar divergence measure and fits a Gaussian mixture model (GMM) on the normalized CE values for partitioning. In contrast, we opt to employ JSD-based selection since it does not require normalization and probabilistic modelling. Besides, unlike CE loss, JSD is symmetric by design and the value ranges from 0 to 1. 

\begin{algorithm}[htb]
	
	\DontPrintSemicolon
	\small
	\textbf{Input:} training set $\mathbb{D} = (\mathcal{X},\mathcal{Y})$, number of samples $N$, number of classes $C$ \\
	    \For{$i=1$ \KwTo  $N$}
	    {
    	    $\mathbf{\mathbf{p}}_i = \big(\mathbf{\Hat{y}}^{(1)}_i+\mathbf{\Hat{y}}^{(2)}_i\big)/2$ \\
    	   $d_i = JSD(\mathbf{p}_i,\mathbf{y}_i)$ (see Eq. \eqref{eq:JSD}) \\
	    }
	   \text{Determine the cutoff distance, $d_{cutoff}$ using Eq.} \eqref{eq:filter_rate} \\
	   $\mathbf{d}_{R} \leftarrow  \{d_i < d_{cutoff}: i \in (1, \dots , N)\} $ \\
	   \text{Determine filter rate,} $R = |\mathbf{d}_{R}|/N $ \\ 
	   $\mathbb{D}_{clean} = \{ \}$ \\
	   \tcp{\footnotesize Uniform Selection} 
	   \For{$j=1$ \KwTo $C$}{
	    $\mathbf{d}^{(j)}_{filtered} \leftarrow  \text{Lowest} \hspace{1mm} R \hspace{1mm} \text{portion of} \hspace{1mm} \mathbf{d}^{(j)} $ \\
	    $\mathbb{D}^{(j)}_{clean} \leftarrow  \{ (\mathbf{x}^{(j)}_t,\mathbf{y}^{(j)}_t): \forall \hspace{0.2cm} d^{(j)}_t \in \mathbf{d}^{(j)}_{filtered} \} $ \\
	    $\mathbb{D}_{clean} \leftarrow   \mathbb{D}_{clean} \cup \mathbb{D}^{(j)}_{clean}$ \\
	    }
	    $\mathbb{D}_{noisy} \leftarrow  \mathbb{D} \setminus \mathbb{D}_{clean}$ \\
		
	\textbf{Output:} $\mathbb{D}_{noisy}, \mathbb{D}_{clean}$
	\caption{\small Uniform Clean Sample Selection}
	\label{alg:UniCon}
\end{algorithm}

After measuring the divergence, $\mathbf{d} =\{d_i: i\in(1, \ldots, N) \}$, for all the samples, we compute a cutoff divergence value, $d_{cutoff}$, which can be expressed as,  
\begin{equation}\label{eq:filter_rate}
  d_{cutoff}  =\left\{
  \begin{array}{@{}ll@{}}
     d_{avg} - {(d_{avg}- d_{min})}/{\tau}, & \text{if} \hspace{0.05cm}  d_{avg} \geq d_{\mu} \\
    d_{avg}, & \hspace{0.05cm} \text{otherwise}
  \end{array}\right.
\end{equation}
where $d_{avg}$ is the average over all values in $\mathbf{d}$, 
$d_{min}$ is the lowest divergence score, $\tau$ is the filter coefficient, and $d_{\mu}$ is an adjustment threshold. Finally, we determine $R$ as the percentage of samples that have JSDs lower than $d_{cutoff}$.  

There are two major benefits of this particular design of $d_{cutoff}$. First, we determine the value of $d_{cutoff}$ based on the network prediction scores (as JSD depends on prediction probabilities) which eliminates the requirement of manual per-dataset tuning. The second benefit stems from the same source, i.e., $d_{cutoff}$ is determined from prediction scores. This ensures that if the network prediction scores are consistently low (high $d_{avg}$), $d_{cutoff}$ will encourage a conservative selection of $\mathbb{D}_{clean}$; which helps in avoiding noisy sample selection at the early stage of training. 


In the next step, we create class-specific partitions, $\{\mathbf{d}^{(1)},\mathbf{d}^{(2)}, \ldots,\mathbf{d}^{(C)}$\}, where $\mathbf{d}^{(j)}$ indicates the JSDs for class $j$. Motivated by the small-loss criterion~\cite{li2020dividemix}, we define \textsc{UniCon} selection criterion as follows:

\textbf{\textsc{UniCon} Selection Criterion:} 
{\it For each class $j$, if the difference $\mathbf{d}^{(j)}_i$ falls within the \underline{lowest $R$ portion} of all values in $\mathbf{d}^{(j)}$, we consider $\mathbf{x}^{(j)}_i$ to have a clean label.}

\noindent Here, $i \in \{1,2, \dots, N_j\}$, $N_j$ is the total number of samples in class $j$, and $\mathbf{x}^{(j)}_i$ is the $i$-th image belonging to the $j$-th class with JSD of $\mathbf{d}^{(j)}_i$. 

Finally, following the \textsc{UniCon} selection criterion, we aggregate all the selected clean and noisy samples from each class to form $\mathbb{D}_{clean}$ and $\mathbb{D}_{noisy}$ with cardinalities of $NR$ and $N(1-R)$, respectively. In cases where the total number of available samples (both clean and noisy) for any class falls below $NR/C$, we take all the available samples in that class for $\mathbb{D}_{clean}$. Algorithm \ref{alg:UniCon} summarizes our selection method.        
\begin{figure}[t]
\vspace{2mm}
\centering
\includegraphics[width=0.55\linewidth]{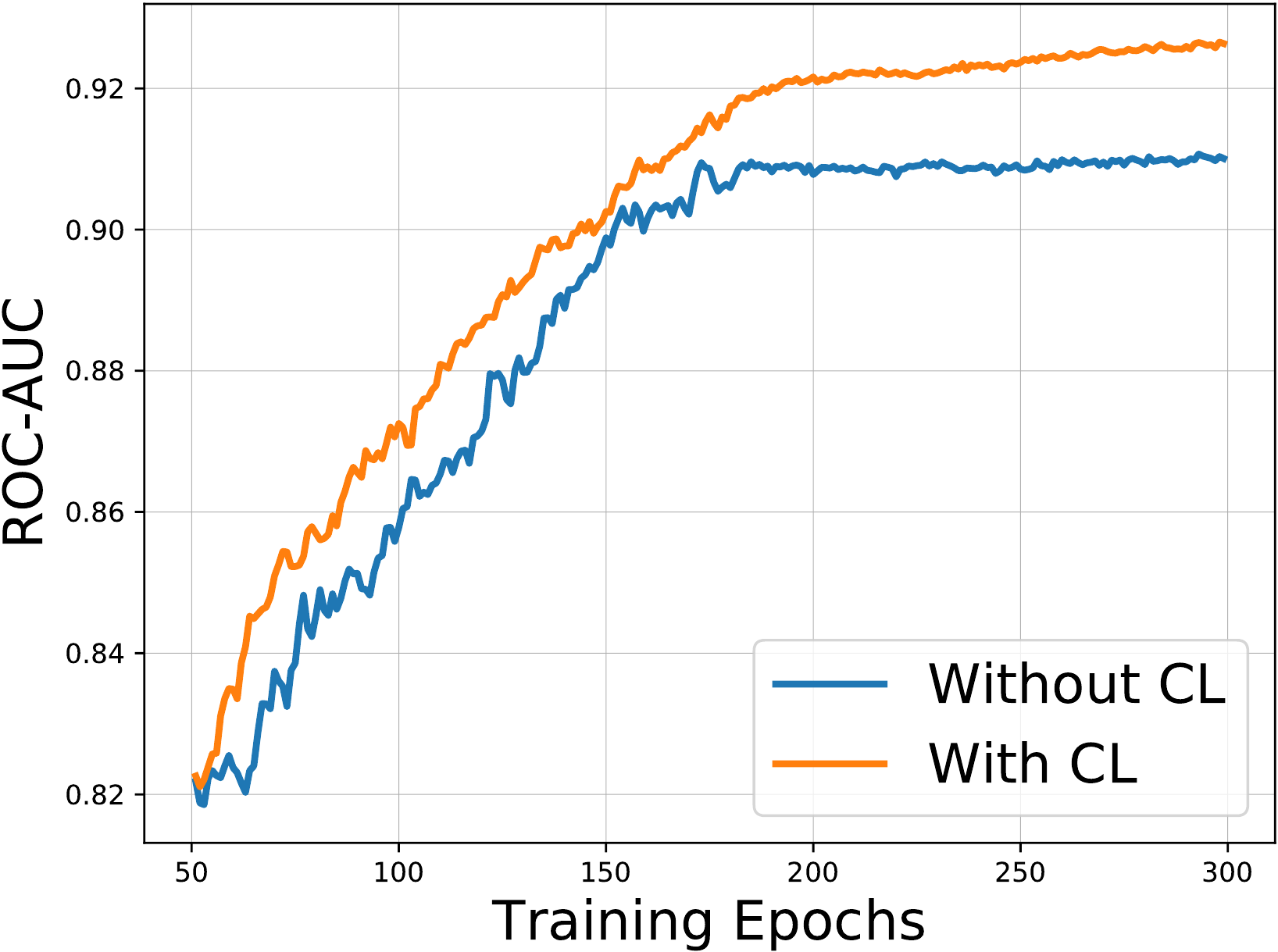}
\vspace{-2mm}
\caption{ROC-AUC score of clean sample selection with and without contrastive learning. As CL helps preventing the memorization, the clean samples are being detected with better precision. Here we considred CIFAR10 with 90\% noise rate. 
}
\label{fig:precision_con}
\vspace{-1mm}
\end{figure}
Note that a previous technique named Jo-SRC \cite{yao2021jo} has employed JSD for clean sample detection. However, our sample selection process differs significantly from Jo-SRC. For instance, the selection threshold in \cite{yao2021jo} needs to be manually fine-tuned during different epochs of the training while \textsc{UniCon} automatically adjusts the filter rate, $R$, based on the network prediction scores; making our proposed selection method hyperparameter independent. 
\subsection{SSL-Training}
\label{sec:ssl_train}
Fig. \ref{fig:sep} shows the details of our SSL-Training with semi-supervised and contrastive loss. Following FixMatch \cite{sohn2020fixmatch}, we perform semi-supervised learning with the samples from $\mathbb{D}_{noisy}$. To this end, we generate two copies of each sample with a weak and a strong augmentation. Pseudo-labels are generated from the weakly augmented copy for computing a semi-supervised loss, $\mathcal{L}_{semi}$, on the strongly augmented copy. We also apply MixUp~\cite{zhang2017mixup} augmentation between the samples from $\mathbb{D}_{clean}$ and $\mathbb{D}_{noisy}$; for the $\mathbb{D}_{noisy}$ samples, we use the pseudo-labels obtained from weakly augmented copy. However, feature or representation learning in such a SSL manner still bears the risk of noise memorization. During training, DNNs memorize certain portion of noisy samples irrespective of the sample selection technique. The presence of such noisy samples in the clean subset, will lead to noisy SSL training. To address this issue, we incorporate \emph{contrastive learning} (CL) \cite{chen2020simple, khosla2020supervised} into our SSL training pipeline to facilitate feature learning without relying on labels/pseudo-labels. Such an unsupervised feature learning scheme further mitigates the risk of noisy label memorization since it does not rely on imperfect separation of clean and noisy samples as well as incorrect pseudo-labels generated during SSL training. Thus, incorporation of CL improves the performance of our proposed selection technique, as shown by the area under the curve (AUC) of Receiver Operating Characteristics (ROC) in Fig. \ref{fig:precision_con}. In our work, we employ contrastive loss only for the samples in the unlabeled set, $\mathbb{D}_{noisy}$. To this end, we employ the projection head $\mathbf{g}(.;\psi)$ to obtain feature projections $\mathbf{z}_{i} = \mathbf{g}(\mathbf{f}({\mathbf{x}}_{i,1}; \theta); \psi)$, and $\mathbf{z}_{j} = \mathbf{g}(\mathbf{f}({\mathbf{x}}_{i,2}; \theta); \psi)$ of the differently augmented copies (${\mathbf{x}}_{i,1}$, ${\mathbf{x}}_{i,2}$) of input  ${\mathbf{x}}_{i}$. The contrastive loss function \cite{chen2020simple,khosla2020supervised} can be expressed as
\begin{align}
        \ell_{i,j}  & = -\log \frac{\exp(\mathrm{sim}(\mathbf{z}_{i}, \mathbf{z}_{j})/\kappa)}{\sum_{b=1}^{2B} \mathbbm{1}_{b \neq i}\exp(\mathrm{sim}(\mathbf{z}_i, \mathbf{z}_b)/\kappa)}~, \\
        & \mathcal{L}_\mathcal{C}= \frac{1}{2B}\sum_{b=1}^{2B}[\ell_{2b-1,2b} + \ell_{2b,2b-1}] \label{eq:loss},
\end{align}
where $\mathbbm{1}_{b \neq i}$ is an indicator function that gives a 1 iff $b \neq i$, $\kappa$ is a temperature constant, $B$ is the number of samples in mini-batch, and $\mathrm{sim}(\mathbf{z}_{i}, \mathbf{z}_{j})$ can be expressed as the cosine similarity between $\mathbf{z}_{i}$ and $\mathbf{z}_{j}$. The total loss function we minimize is
\begin{equation} \label{eq:main_loss}
    \mathcal{L}_{tot} = \mathcal{L}_{semi} + \lambda_{\mathcal{C}} \mathcal{L}_\mathcal{C},
\end{equation}
where $\lambda_{\mathcal{C}}$ is contrastive loss coefficient. Additional details of the contrastive learning as well as the rest of our SSL-Training scheme is provided in the {\em supplementary material}.

\section{Experimental Settings}
\subsection{Datasets}
\textbf{CIFAR10/100:} The CIFAR-10/100 datasets \cite{krizhevsky2009learning} contain 50K training and 10K test images.
In general, it is difficult to control or determine the noise characteristics; e.g. noise rate, in natural datasets. Therefore, synthetic noise models are commonly used for the evaluation of noise-robust algorithms. In our work, we employ two types of noise models: symmetric and asymmetric. For symmetric noise model, an $r$ portion of samples from one particular class are uniformly distributed to all other classes.
On the other hand, the design of asymmetric label noise follows the structure of real mistakes that take place in CIFAR10 \cite{li2019learning}: “Truck→ Automobile, Bird → Airplane, Deer → Horse, Cat → Dog”.  For CIFAR100, we use label flips for each class to the next one within the super-classes.  


\textbf{Tiny-ImageNet \cite{le2015tiny}:} This dataset is a smaller version of the original ImageNet in terms of the number of classes and the image resolution. There are in total 200 classes containing 500 images per class. The image size is $64 \times 64$. 

\textbf{Clothing1M:} Clothing1M is a large-scale
real-world dataset with noisy labels \cite{xiao2015learning}. It contains 1M images from 14 different cloth-related classes.
Since the labels are produced by the seller provided surrounding texts of the images, a large portion of confusing classes (e.g., Knitwear and Sweater) are mislabeled. 

\textbf{Webvision~\cite{li2017webvision}:} This dataset contains 2.4 million images (obtained from Flickr and Google)
that are categorized into the same 1,000 classes as in the ImageNet ILSVRC12. 
Following the previous studies \cite{li2020dividemix, nishi2021augmentation}, we use the first 50 classes of the Google image subset as the training data. 

\subsection{Training Details}
We use the PreAct ResNet18~\cite{he2016deep} architecture for CIFAR10, CIFAR100, and Tiny-ImageNet. For Clothing1M and WebVision datasets, we take a ResNet50 network~\cite{he2016deep} pre-trained on ImageNet and a InceptionResNetV2 network \cite{szegedy2017inception} which is trained from scratch. We modify these architectures with a projection head, that produces a embedding vector of size 128,  to facilitate contrastive learning.

For CIFAR-10 and CIFAR-100, optimization is performed using stochastic gradient descent (SGD) optimizer with the following settings:  an initial learning rate (LR) of $0.02$, a weight decay of $5e^{-4}$, a value of 0.9 for the momentum, and a batch size of $64$. For CIFAR-10 and CIFAR-100, we train each network for around 300 epochs while linearly decaying the learning rate (lr-decay) by 0.1 per 120 epochs. 
Following \cite{li2020dividemix}, a warmup period of 10 and 30 epochs was employed before starting the selection and SSL-Training. 
For Tiny-ImageNet, we use an initial LR of 0.01, a weight decay of $1e^{-3}$ with a batch size of $32$. We train the network for 350 epochs and the lr-decay rate is 0.1/100 epochs. The warmup period is 15 epochs. 
For Clothing1M, we choose an initial LR of 0.002 and a weight decay of $1e^{-3}$. We employ the same settings as Tiny-ImageNet for WebVision. The total number of training epochs is 100 and the lr-decay rate is 0.1/40 epochs. 
 
For data augmentations, we follow Auto-augment policy described in \cite{cubuk2019autoaugment}. For CIFAR-10 and CIFAR100, we use CIFAR10-Policy and we apply ImageNet-Policy to Tiny-ImageNet. As these policies are transferable from one dataset to another, ImageNet-Policy is employed for both Clothing1M and Webvision dataset.

\begin{table}[t]
\centering
\scalebox{0.78}{
\begin{tabular}{l|cccc|cccc} \toprule
&  \multicolumn{4}{c|}{\textbf{CIFAR-10}}  &  \multicolumn{4}{c}{\textbf{CIFAR-100}}  \\  \cline{2-9}  
Method  &  20\%  &  50\%  &  80\%  &  90\%  &  20\%  &  50\%  &  80\%  &  90\%  \\  \midrule
  
{CE}   &  86.8  &  79.4  &  62.9  &  42.7  &  62.0  &  46.7  &  19.9  &  10.1  \\ 
{LDMI \cite{xu2019ldmi}}   &  88.3  &  81.2  &  43.7  &  36.9 &   58.8  &  51.8  &  27.9  &  13.7 \\ 
{M-Up \cite{zhang2018mixup}}   &  95.6  &  87.1  &  71.6  &  52.2 &  67.8  &  57.3  &  30.8  &  14.6 \\ 
{PCIL \cite{yi2019probabilistic}}  &  92.4  &  89.1  &  77.5  &  58.9  &  69.4  &  57.5  &  31.1  &  15.3  \\ 
{JPL \cite{kim2021joint}} &  93.5  &  90.2  &  35.7  &  23.4 &   70.9   &  67.7  &  17.8  &  12.8  \\ 
{MOIT \cite{ortego2021multiobjective}}  &  94.1  &  91.1  &  75.8  &  70.1 &  75.9  &  70.1  &  51.4  &  24.5  \\ 
{DMix \cite{li2020dividemix}}   &  \textbf{96.1}  &  94.6  &  92.9  &  76.0   &  77.3  &  74.6  &  60.2  &  31.5  \\ 
{ELR \cite{liu2020earlylearning}}  &  95.8  &  94.8  &  93.3  &  78.7  &  77.6  &  73.6  &  60.8  &  33.4  \\ \midrule

\textsc{UniCon}  &  96.0  &  \textbf{95.6}  &  \textbf{93.9}  &  \textbf{90.8}   &   \textbf{78.9}  &  \textbf{77.6}  &  \textbf{63.9}  &  \textbf{44.8}  \\   \bottomrule
\end{tabular}
}
\vspace{-1mm}
\caption{Test accuracies (\%) obtained by different techniques under symmetric noise. Our class balance with contrastive loss strategy improves performance at almost every noise level. Results for previous techniques were copied from their respective papers.}
\label{tab:symmetry_cifar10_100}
\vspace{-3mm}
\end{table}

\begin{table}[t]
\vspace{-2.5mm}
\centering
\medskip{}
\scalebox{0.78}{
\begin{tabular}{l|ccc|ccc}\toprule
&  \multicolumn{3}{c|}{\textbf{CIFAR-10}}  &  \multicolumn{3}{c}{\textbf{CIFAR-100}}  \\  \cline{2-7}  
Method  &  10\%  &  30\%  &  40\%   &  10\%  &  30\%  &  40\%   \\  \midrule
  
{CE}   &  88.8  &  81.7  &  76.1  &  68.1  &  53.3  &  44.5  \\ 
{LDMI \cite{xu2019ldmi}}   &  91.1  &  91.2  &  84.0  &  68.1  &  54.1  &  46.2   \\ 
{M-Up \cite{zhang2018mixup}}   &  93.3  &  83.3  &  77.7  &  72.4  &  57.6  &  48.1   \\ 
{JPL \cite{kim2021joint}} &  94.2  &  92.5  &  90.7  &  72.0  &  68.1  &  59.5  \\ 
{PCIL \cite{yi2019probabilistic}}  &  93.1  &  92.9  &  91.6  &  76.0  &  59.3  &  48.3  \\ 
{DMix$^{*}$ \cite{li2020dividemix}}  &  93.8  &  92.5  &  91.7  &  71.6  &  69.5  &  55.1  \\ 
{ELR$^{*}$ \cite{liu2020earlylearning}}  &  \textbf{95.4}  &  94.7  &  93.0 &  77.3  &  74.6  &  73.2   \\ 
{MOIT \cite{ortego2021multiobjective}}  &  94.2  &  94.1  &  93.2  &  77.4  &  75.1  &  74.0  \\ \midrule
\textsc{UniCon}  &  95.3  &  \textbf{94.8}  &  \textbf{94.1}  &  \textbf{78.2}  &  \textbf{75.6}  &  \textbf{74.8}  \\ \bottomrule

\end{tabular}
}
\vspace{-1mm}
\caption{Experimental results on CIFAR10 and CIFAR100 with asymmetric noise. \textsc{UniCon} sees consistent improvement for CIFAR100 dataset under different asymmetric noise settings. $(*)$ indicates that we run the algorithm.}
\label{tab:asymmetry_cifar10_100}
\vspace{-2mm}
\end{table}

\section{Experimental Results}
We present the performance of \textsc{UniCon} 
under different label noise scenarios. We start with the synthetic noisy label datasets (e.g. CIFAR10, CIFAR100 and TinyImageNet) and move on to the real world noisy datasets (e.g. WebVision, Clothing1M). For experiments, we consider symmetric noise rates of 20\%, 50\%, 80\%, and 90\% and asymmetric noise rates of 10\%, 30\%, and 40\%. 

\textbf{CIFAR10 and CIFAR100 datasets:}  Table \ref{tab:symmetry_cifar10_100} shows the average test accuracies for these datasets. In case of CIFAR10, from moderate to severe label noise, \textsc{UniCon} performs consistently better than the baseline methods. 
For $90\%$ noise rate, we achieve a significantly better performance improvement over the state-of-the art. For high noise rate, techniques like \cite{li2020dividemix} usually fail due to high number of false positives. 
However, for low noise rate (20\%), \cite{li2020dividemix} performs slightly better than ours. Low noise rate indicates more clean samples are available for supervised learning. One possible explanation could be that the scarcity of unlabeled data (i.e. $|\mathbb{D}_{noisy}| < |\mathbb{D}_{clean}|$) makes contrastive feature learning less effective.  We have also conducted experiments under the asymmetric noise scenario. In case of asymmetric noise, each class is not equally affected by label noise. This makes the selection of clean samples a bit more challenging. However, \textsc{UniCon} achieves similar performance gain as symmetric noise which is shown in Table \ref{tab:asymmetry_cifar10_100}. Note that there is an exception at 10\% noise rate as \cite{liu2020earlylearning} obtains $0.1\%$ better accuracy than \textsc{UniCon}. 

\begin{table}[htb]
\vspace{-1.5mm}
\centering
\medskip{}
\scalebox{0.87}{
\begin{tabular}{l|cc|cc|cc} \toprule
Noise (\%) & \multicolumn{2}{c|}{0} & \multicolumn{2}{c|}{20} & \multicolumn{2}{c}{50} \\  \midrule
Alg. & Best & Avg. & Best & Avg. & Best & Avg.  \\ \midrule

Standard CE & 57.4 & 56.7 & 35.8 & 35.6 & 19.8 & 19.6  \\
{Decoupling \cite{malach2017decoupling}} & - & - & 37.0 & 36.3 & 22.8 & 22.6  \\
{F-correction \cite{patrini2016making}} & - & - & 44.5 & 44.4 & 33.1 & 32.8  \\
{MentorNet \cite{jiang2018mentornet}} & - & - & 45.7 & 45.5 & 35.8 & 35.5 \\
{Co-teaching+ \cite{yu2019does}} & 52.4 & 52.1 & 48.2 & 47.7 & 41.8 & 41.2 \\
{M-correction \cite{arazo2019unsupervised}} & 57.7 & 57.2 & 57.2 & 56.6 & 51.6 & 51.3 \\
{NCT \cite{sarfraz2020noisy}} & 62.4 & 61.5 & 58.0 & 57.2 & 47.8 & 47.4 \\ \midrule

\textsc{UniCon} & \textbf{63.1}& \textbf{62.7} & \textbf{59.2} & \textbf{58.4} & \textbf{52.7} & \textbf{52.4} \\\bottomrule

\end{tabular}
}
\caption{Test accuracies (\%) on Tiny-ImageNet dataset under symmetric noise settings. We report the results for other methods directly from \cite{sarfraz2020noisy} with the highest (Best) and the average (Avg.) test accuracy (\%) over the last 10 epochs.}
\label{tab:tiny-imagenet}
\end{table}
Table \ref{tab:symmetry_cifar10_100} and \ref{tab:asymmetry_cifar10_100} contain the average test accuracies for CIFAR100 dataset. 
\textsc{UniCon} shows similar effectiveness against label noise in CIFAR100 obtaining an accuracy improvement of 11.4\%  for 90\% noise rate. This improvement is consistent under different noise settings. 
While  ELR \cite{liu2020earlylearning}, DMix \cite{ li2020dividemix} and MOIT \cite{ortego2021multiobjective} show some level of resistance to noisy labels for low noise rate, the performances are not consistent for high noise rate. Furthermore, the asymmetric noise performance of our method are also superior than other baseline methods in Table \ref{tab:asymmetry_cifar10_100}. 

\begin{table}[t]
\centering
\scalebox{0.9}{
\begin{tabular}{l|l|c} \toprule
Method & Backbone & Test Accuracy\\ \midrule
Standard CE  & ResNet-50 &  69.21 \\
Joint-Optim \cite{tanaka2018joint} & ResNet-50 & 72.00 \\
MetaCleaner \cite{zhang2019metacleaner} & ResNet-50 & 72.50\\
MLNT \cite{li2019learning}     & ResNet-50 &   73.47    \\ 
PCIL \cite{yi2019probabilistic} & ResNet-50 & 73.49  \\
JPL \cite{kim2021joint}  & ResNet-50 &  74.15   \\
DMix \cite{li2020dividemix}  & ResNet-50 & 74.76   \\
ELR \cite{liu2020earlylearning} & ResNet-50 & 74.81   \\\midrule
\textsc{UniCon} & ResNet-50 & \textbf{74.98} \\\bottomrule
\end{tabular}
}
\caption{Experimental results on Clothing1M dataset.  Results for previous techniques were copied from their respective papers. }
\label{tab:clo1m}
\end{table}

\begin{table}[t]
\centering
\scalebox{0.9}{
\begin{tabular}{l|cc|cc} \toprule
Dataset & \multicolumn{2}{c}{WebVision} & \multicolumn{2}{|c}{ ILSVRC12}\\ \midrule
Method & \emph{Top-1} & \emph{Top-5} & \emph{Top-1} & \emph{Top-5} \\ \midrule
D2L \cite{ma2018dimensionality} & 62.68 & 84.00 & 57.80 & 81.36 \\
MentrorNet \cite{jiang2018mentornet} & 63.00 & 81.40 & 57.80 & 79.92 \\
Co-Teaching \cite{han2018co} & 63.58 & 85.20 & 61.48 & 84.70 \\
Iterative-CV \cite{wang2018iterative} & 65.24 & 85.34 & 61.60 & 84.98 \\
DivideMix \cite{li2020dividemix} &  77.32 & 91.64 & 75.20 & 90.84 \\
ELR \cite{liu2020earlylearning} & 77.78  & 91.68 & 70.29 & 89.76\\ 
MOIT \cite{ortego2021multiobjective} & \textbf{78.76} & -  & - & -\\ \midrule
\textsc{UniCon} & 77.60 & \textbf{93.44} & \textbf{75.29} & \textbf{93.72}  \\ \bottomrule
      
\end{tabular}
}
\caption{\small Experimental results on Webvision and ILSVRC12. All methods are trained on the Webvision while evaluated on both Webvsion and ILSVRC12 validation set. Results for baseline methods are taken from \cite{liu2020earlylearning} and \cite{ortego2021multiobjective}. MOIT \cite{ortego2021multiobjective} does not evaluate their method on ILSVRC12 and did not provide top-5 accuracies.
}
\label{tab:webvision}
\vspace{-2mm}
\end{table}

\begin{figure}[t]
  \centering
  \begin{subfigure}{0.475\linewidth}
    \includegraphics[width=0.98\linewidth]{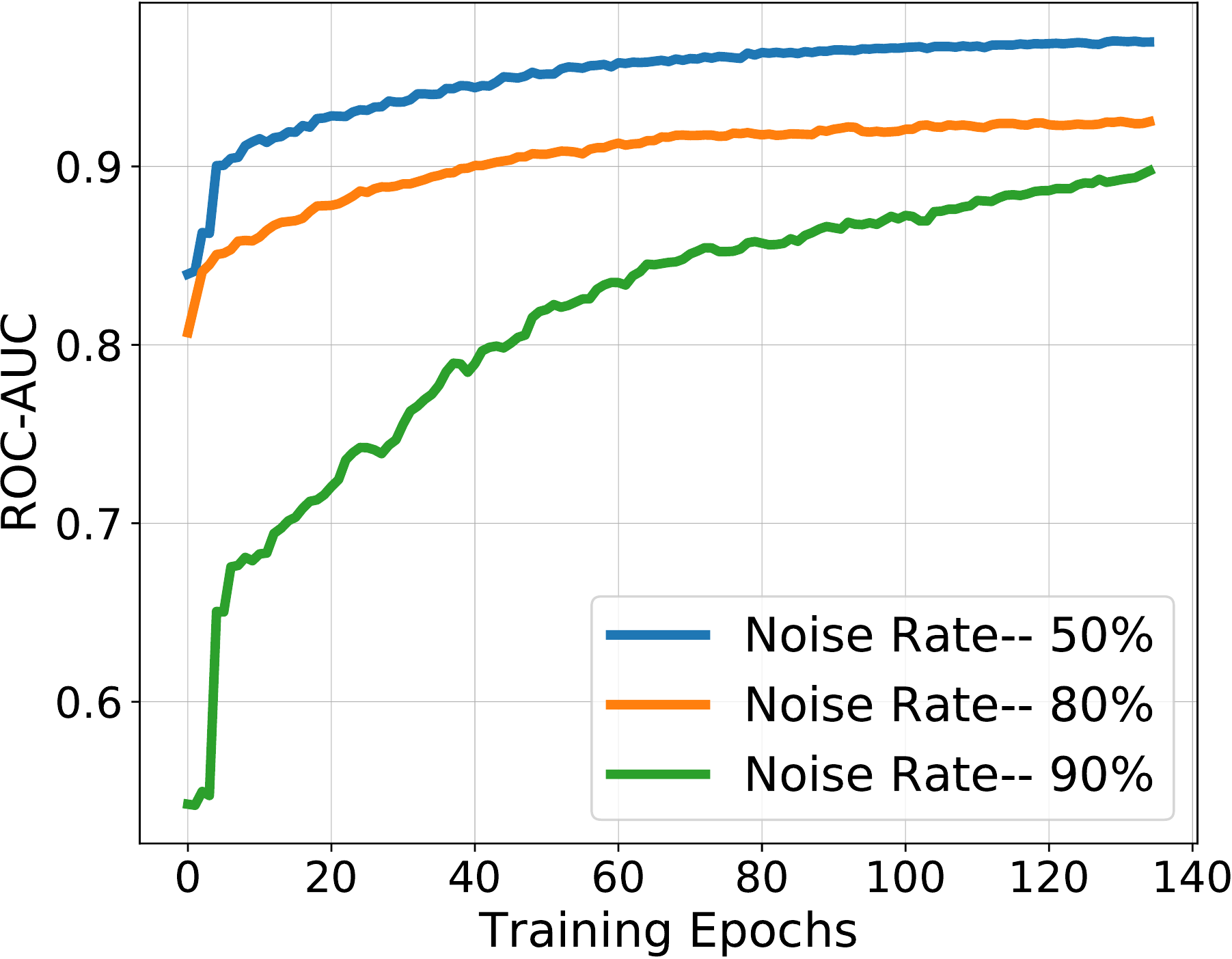}
    \caption{ROC-AUC score vs epochs.}
    \label{fig:ROC_AUC}
  \end{subfigure}
  \hfill
  \begin{subfigure}{0.475\linewidth}
    \includegraphics[width=0.98\linewidth]{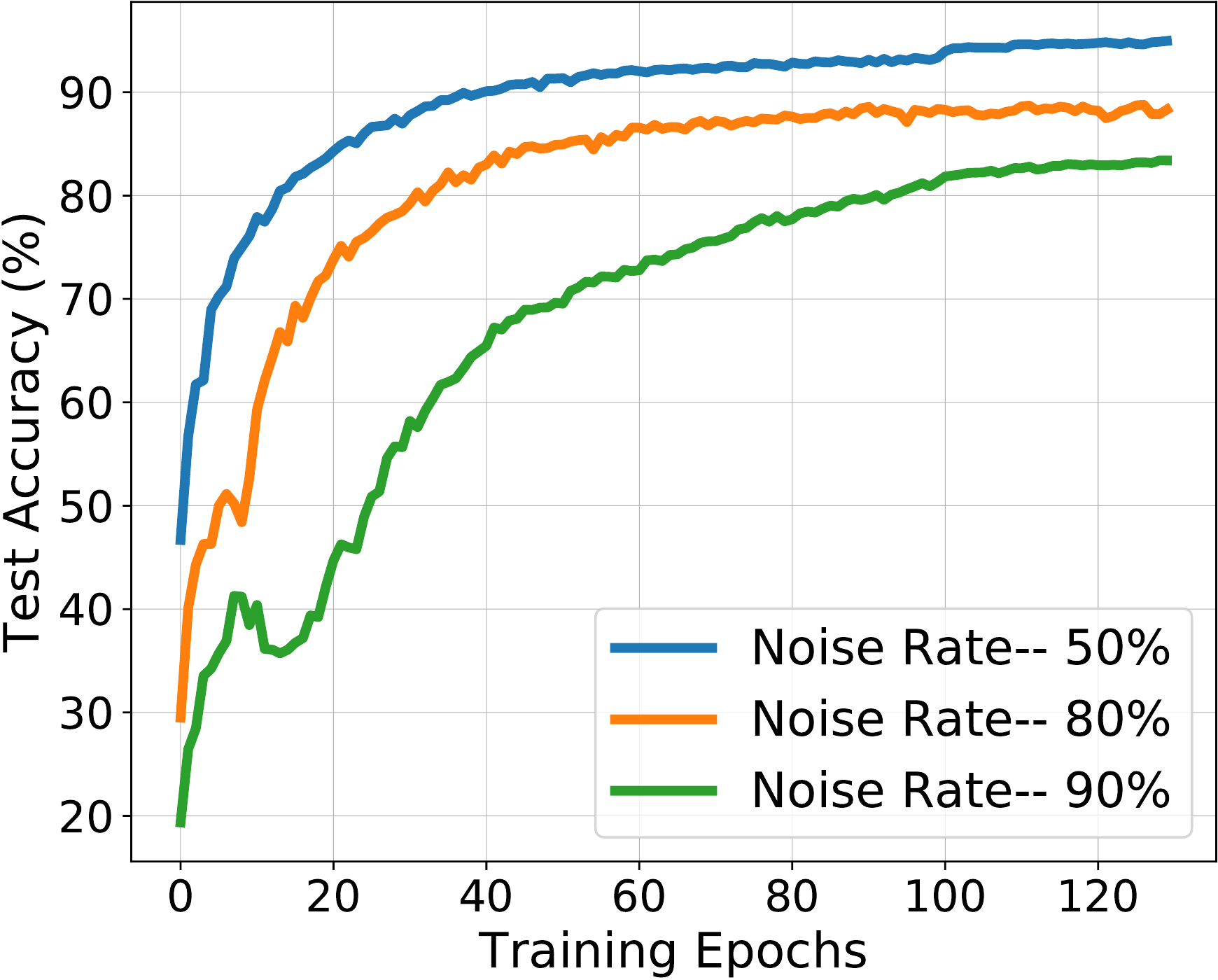}
    \caption{Test Accuracy vs Epochs}
    \label{fig:test_acc}
  \end{subfigure}
  \vspace{-1mm}
  \caption{ROC-AUC score and test accuracy ($\%$) on CIFAR10 with different noise rates. As the model becomes more precise in selection, the test-time performance improves accordingly.}
  \label{fig:ROC_accuracy}
  \vspace{-2mm}
\end{figure}
\begin{table*}[htb]
\centering
\scalebox{0.9}{
\begin{tabular}{l|cccccc|cccccc} \toprule
Dataset & \multicolumn{6}{c}{CIFAR10} & \multicolumn{6}{|c}{CIFAR100} \\ \midrule
Noise Rate & \multicolumn{2}{c}{50\%} & \multicolumn{2}{|c}{80\%} & \multicolumn{2}{|c}{90\%} & \multicolumn{2}{|c}{50\%} & \multicolumn{2}{|c}{80\%} & \multicolumn{2}{|c}{90\%} \\ \midrule
Method & Best & Last & Best & Last & Best & Last & Best & Last & Best & Last & Best & Last\\ \midrule
\textsc{UniCon} w/o balancing & 94.28 & 94.06 & 91.41 & 91.16 & 85.49 & 85.28  & 75.26 & 75.01 & 60.51 & 60.16 & 39.87 & 39.02 \\
\textsc{UniCon} w/o CL & 94.92 & 94.24 & 91.67 & 91.21 & 87.28 & 86.34 & 75.75 & 75.09 & 60.54 & 60.17 & 41.83 & 41.11 \\

\textsc{UniCon} w/o ensemble & 95.20 & 94.91  & 92.38 & 92.11 & 88.84 & 88.18 & 76.28 & 76.10  & 62.98 & 62.11 & 42.36 & 41.56 \\ \midrule
\textsc{UniCon}  & \textbf{95.61} & \textbf{95.24} & \textbf{93.97} & \textbf{93.97} & \textbf{90.81}& \textbf{89.95}& \textbf{77.63} & \textbf{76.91} & \textbf{63.98} & \textbf{63.13} & \textbf{44.82} & \textbf{44.51} \\ \bottomrule
\end{tabular}
}
\vspace{-2mm}
\caption{Ablation study with different training settings. Both contrastive loss and class-imbalance affects the performance significantly; especially for high noise rates. Ensembling the outputs of both network during separation seems to improve the performance as well. Test results at last epoch are also shown here.}
\label{tab:ablation_con_imb_ens}
\vspace{-0.5cm}
\end{table*}
\begin{figure}
    \centering
    \includegraphics[width=0.55\linewidth]{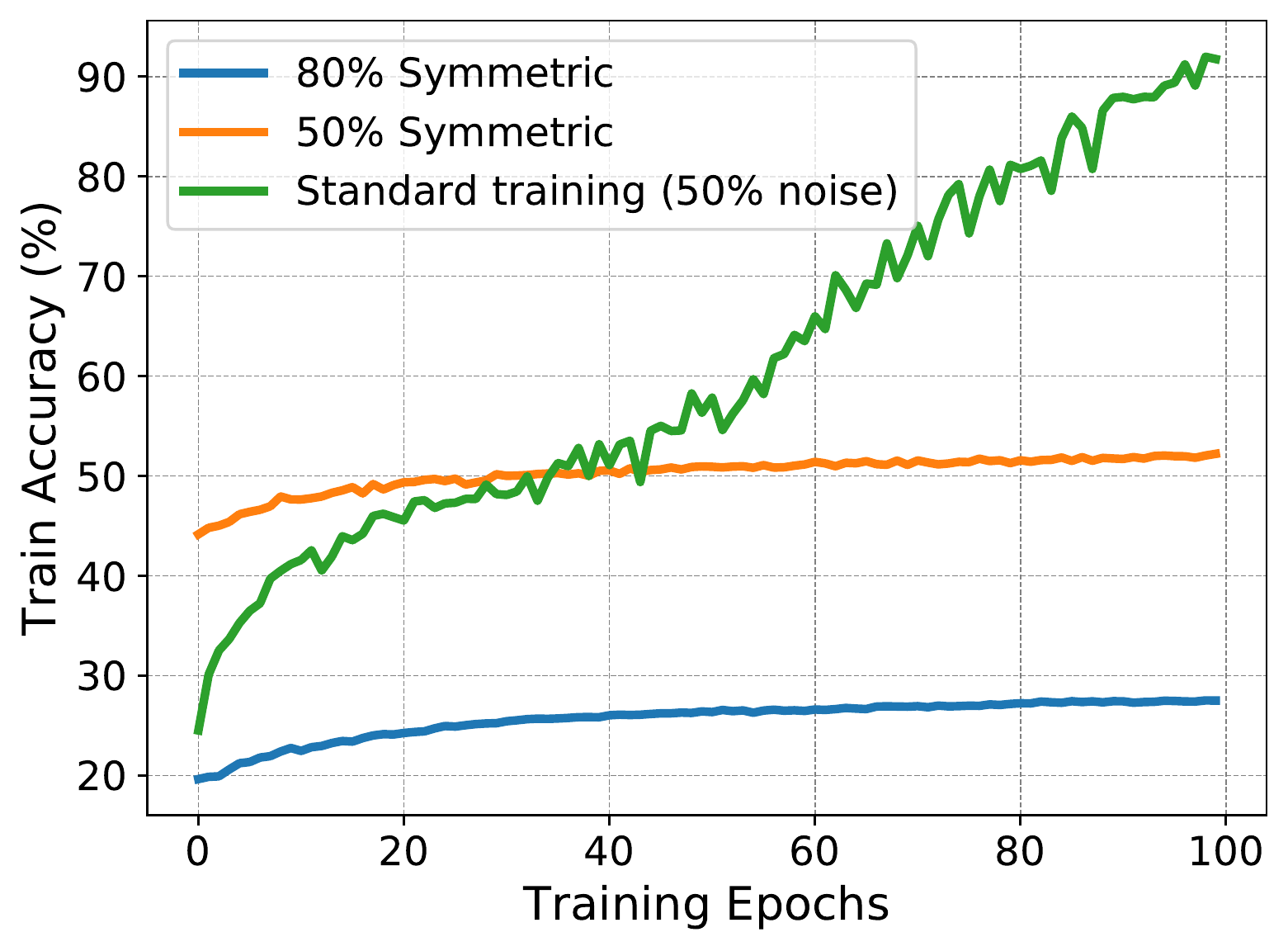}
    \vspace{-2mm}
    \caption{Training accuracy at different epochs. Low accuracy indicates that the networks do not memorize the noisy labels even after long training. In contrast to UNICON, standard CE loss-based training leads to a high training accuracy (should be $\sim$50\%), i.e., complete memorization of noisy labels.}
    \label{fig:memorization}
    \vspace{-2mm}
\end{figure}
 \textbf{TinyImageNet Dataset:} Table \ref{tab:tiny-imagenet} presents the performance comparison of \textsc{UniCon} and other state of the art methods. 
Even with no label noise, Tiny-ImageNet remains a challenging benchmark dataset to deal with. 
It becomes more challenging under the presence of label noise. One of the baseline methods M-correction \cite{arazo2019unsupervised} uses a loss-correction technique to tackle noisy labels while NCT \cite{sarfraz2020noisy} leverages from collaborative learning of two networks. However, both methods underperform compared to our method. Table \ref{tab:tiny-imagenet} shows that \textsc{UniCon} gains around $1\%$ performance improvement over SOTA for all noise rates.

 \textbf{Clothing1M Dataset:} 
Table \ref{tab:clo1m} presents performance comparison on this real world noisy labeled dataset. 
We achieve 0.17\% performance improvement over ELR \cite{liu2020earlylearning}. The performance improvement for clothing1M sometimes depends on the length of warmup, as longer period of standard CE-based training can lead to memorization. In our training, we use a warm-up period of 2,000 steps.  

\textbf{WebVision Dataset:} We present our experimental results on this dataset in Table \ref{tab:webvision}. While validating, MOIT \cite{ortego2021multiobjective} sees SOTA \emph{Top-1} accuracy while our method achieves the best  \emph{Top-5} accuracy. We obtain around $1.5\%$ improvement over SOTA (MOIT \cite{ortego2021multiobjective} did not provide \emph{Top-5} accuracy.)  Furthermore, \textsc{UniCon} secures SOTA \emph{Top-1} and \emph{Top-5} accuracies on ILSVRC12 validation set. While the gain in \emph{Top-1} accuracy is not significant, we achieve a performance improvement of $1.88\%$ over DMix \cite{li2020dividemix} in \emph{Top-5} accuracy. 


\subsection{Ablation Studies}
In this section, we conduct an ablation study of \textsc{UniCon} under different training settings.

 \textbf{Sample Selection Performance:}
In general, the precision of clean sample selection directly impacts the overall performance of any selection-based noisy label technique. Likewise, the success of \textsc{UniCon} depends on how well it can separate the clean samples. Fig.~\ref{fig:ROC_AUC} shows the the ROC-AUC score of our selection mechanism under different noise settings. It can be observed that \textsc{UniCon} sees a steady rise in the precision irrespective of the noise level. In case of high noise rate, it is usual for the network to get confused between clean and noisy samples. However, our separation approach proves to be effective even under such scenario. With improved precision, the network learns better discriminative features from labeled data and generalizes well to the unlabeled data. Through the generation of quality pseudo-labels, \textsc{UniCon} improves the classification accuracy significantly (Fig.~\ref{fig:test_acc}).   

 \textbf{Effect of Contrastive Learning:} 
CL is one of the key components of our framework. Table \ref{tab:ablation_con_imb_ens} indicates the impact of CL in overall performance of our method. As CL is resistant to label noise memorization, it boosts the performance significantly even in high label noise scenarios. For CIFAR10 and CIFAR100, with 90\% noise rate, \textsc{UniCon} without CL sees $3.53\%$ and $2.99\%$ drop in test accuracies respectively. We explain more on contrastive learning and its impact in the {\em supplementary material.}


 \textbf{Effect of Ensemble and balancing:}
During selection, we take the average of both network's predictions instead of depending on just one network \cite{han2018co}. This seems to improve the performance significantly in case of high noise rate (see Table \ref{tab:ablation_con_imb_ens}). However, taking the feedback from both networks bears the risk of confirmation bias over the course of training \cite{li2020dividemix}. We prevent that by training one network at a time. During the same training epoch, we perform the separation again before training the other network. Table \ref{tab:ablation_con_imb_ens} also contains the performance of our method without balancing. The significant decrease in classification accuracies underlines the importance of class-balance prior. The effectiveness of \textsc{UNICON} in combating memorization can be observed in Fig.~\ref{fig:memorization}.

\section{Limitations of \textsc{UniCon}}
In this work, to combat label noise we employ a class-balance prior. The prior helps in combating artificial imbalance caused by current state-of-the-art selection methods. This prior can be restrictive in some extreme scenarios where the dataset itself exhibits extreme imbalance. However, in such cases, it is possible to update our prior accordingly based on the class distribution of the dataset. Since knowing the dataset distribution in advance is equally restrictive we do not explore this direction in this study. Additionally, even though we provide a general solution for combating label noise, our solution is particularly effective under high label noise. Therefore, it is possible to outperform our proposed method on datasets which do not contain a significant amount of label noise. However, we emphasize that such success can be attributed to superior training strategy and complicated design whereas our simple solution is more general and provides reasonable results even for such low noise rate scenarios.  
\section{Conclusion}
In this work, we proposed \textsc{UniCon}, a simple yet effective solution for combating label noise. Our proposed uniform selection technique effectively addresses often overlooked but critical shortcoming of selection based state-of-the-art methods. Furthermore, our constrastive feature learning approach provides a fundamental solution to combat memorization of noisy label. Equipped with these two components, our method selects clean samples more precisely over the course of training by reducing the class-disparity among the true positives and CL-based unsupervised feature learning. Network trained on high precision clean samples generates higher quality pseudo-labels for the noisy label data and the overall process improves the high noise level performance significantly. \textsc{UniCon} achieves  $\sim$10\% performance improvement over state-of-the-art on 90\% noisy CIFAR10 and CIFAR100. Through extensive empirical analysis, we show the effectiveness of our method under different noise scenarios. 

\textbf{Acknowledgement:} Professor Ajmal Mian is the recipient of an Australian Research Council Future Fellowship Award (project number FT210100268) funded by the Australian Govt. This work is partly supported by the National Science Foundation under Grant No. CCF-1718195. 






\newpage
{\small
\bibliographystyle{ieee_fullname}
\bibliography{egbib}
}
\newpage

\textbf{\Large \centering Supplementary Material}
\section{Overview}
Section \ref{sec:SSL_Training} describes our SSL-Training method in detail. Section \ref{sec:ablation_studies} discusses the findings of our ablation studies. Section \ref{sec:training_details} has some details about hyperparameter settings and experimental results. 

\begin{figure*}[htb]
    \centering
    \includegraphics[width=1\linewidth]{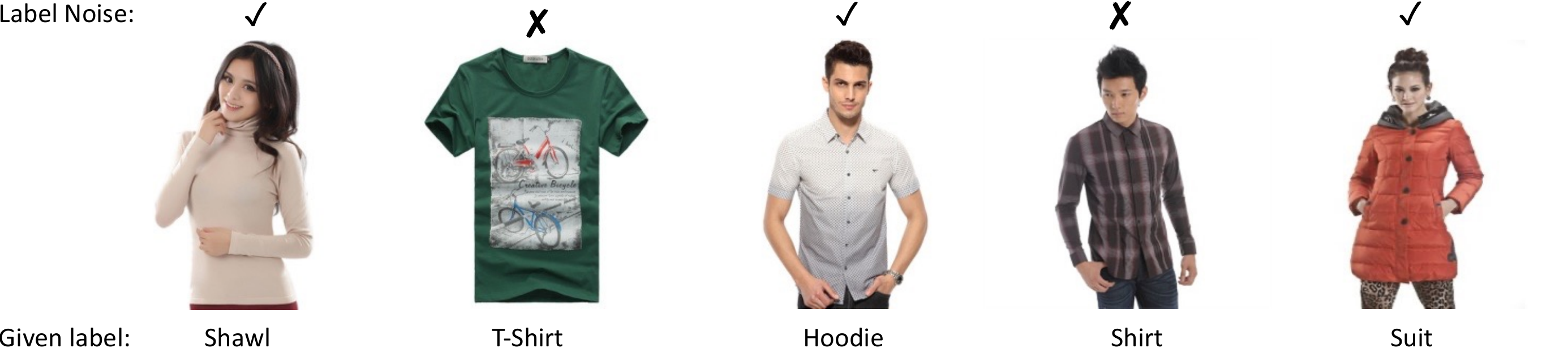}
    \caption{Sample images from Clothing1M~\cite{xiao2015learning} dataset. We show the given label (bottom) and indicate label noise (top) for each image. Noisy samples are marked as positive (red) while clean samples contain negative marks (green). 
    Here, the noise rate is 60\% (3/5). Note that these images are taken for demonstration purpose only and corresponding labels are not their original given labels. 
    }
    \label{fig:examples}
\end{figure*}

\begin{figure*}[htb]
\centering
\includegraphics[width=1\linewidth]{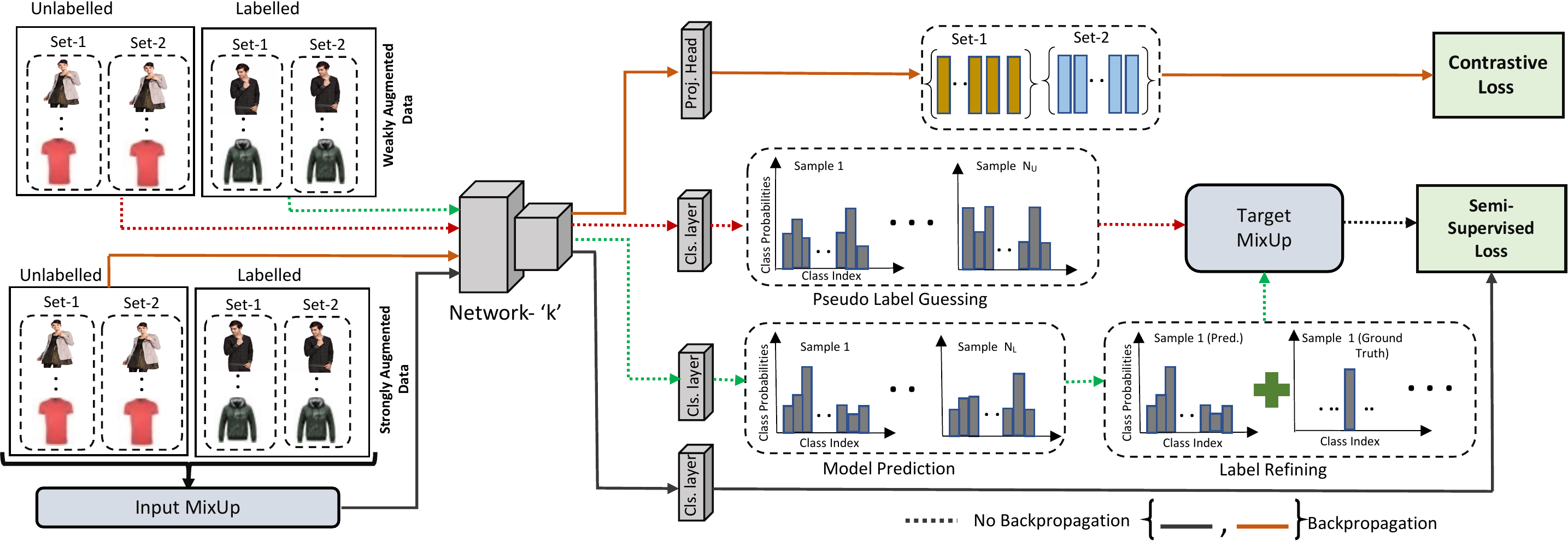}
\caption{SSL-Training of network-`k' (k=1,2). After separating the samples, we create total 8 sets of weak and strong augmented data. While weakly augmented data helps with target label generation, strongly augmented data are used for updating the parameters through backpropagation. There are two types of label generation here: pseudo label guessing (\ref{eq:label_guessing})  (represented by green color) and label-refinement (\ref{eq:label_refinement}) (represented by red color). We have semi-supervised (eq.~\ref{eq:combined_loss}) and contrastive (eq.~\ref{eq:con_loss}) losses that are minimized during training. 
Note that for pseudo-label guessing we take the average of both network-(1,2) predictions which is not shown here (eq. \ref{eq:label_guessing}).}
\label{fig:framework_sup}
\end{figure*}

\section{SSL-Training Details} \label{sec:SSL_Training}
Before semi-supervised learning (SSL), we separate the training set into $\mathbb{D}_{clean}$ and $\mathbb{D}_{noisy}$ by applying uniform selection. A sample training set with $60\%$ noise level is shown in Figure \ref{fig:examples}. We consider $\mathbb{D}_{clean}$ and $\mathbb{D}_{noisy}$ to be labeled and unlabeled data, respectively. At the beginning of SSL, we create four sets of weakly-augmented (WA) data:
\begin{itemize}
    \item Two sets of weakly-augmented labeled data $\{\hat{\mathbf{x}}^{weak}_{i,1}, \hat{\mathbf{x}}^{weak}_{i,2}: i\in(1,...,N)\}$.
    \item Two sets for weakly-augmented unlabeled data $\{\hat{\mathbf{u}}^{weak}_{i,1}, \hat{\mathbf{u}}^{weak}_{i,2}: i\in(1,...,N)\}$.
\end{itemize}
In addition, we also generate four sets of strongly-augmented (SA) data: 
\begin{itemize}
    \item Two sets of strongly-augmented labeled data $\{\hat{\mathbf{x}}^{strong}_{i,1}, \hat{\mathbf{x}}^{strong}_{i,2}: i\in(1,...,N)\}$.
    \item Two sets of strongly-augmented unlabeled $\{\hat{\mathbf{u}}^{strong}_{i,1}, \hat{\mathbf{u}}^{strong}_{i,2}: i\in(1,...,N)\}$.
\end{itemize}

Here, weak augmentations are used for label updating (label-refinement and pseudo-label guessing). We employ strong augmentations for updating the network parameters using backpropagation. For label-refinement \cite{li2020dividemix}, we use the networks' prediction to a weakly-augmented sample $\mathbf{x}_i$ for refining the given-label $\mathbf{y}_i$. For $\{\hat{\mathbf{x}}^{weak}_{i,1}, \hat{\mathbf{x}}^{weak}_{i,2}\}$, the output probabilities can be written as,
\begin{equation} \label{eq:label_re}
    \mathbf{p}_i=\frac{1}{2}\sum_{m=1}^{2}\mathbf{h}(\mathbf{f}(\hat{\mathbf{x}}^{weak}_{i,m};\theta^{(k)}); \phi^{(k)}),
\end{equation}
where $N$ is the number of data points in the training set and $\mathbf{h}(\mathbf{f}(\hat{\mathbf{x}}^{weak}_{i,m};\theta^{(k)}); \phi^{(k)})$ is the Softmax probabilities of  network-$k$ ($k$=1,2) corresponding to $\hat{\mathbf{x}}^{weak}_{i,m}$.

After getting $\mathbf{p}_i$, we refine the label as follows:
\begin{equation} \label{eq:label_refinement}
    \bar{\mathbf{y}}_i=w_i \mathbf{y}_i+(1-w_i)\mathbf{p}_{i}, 
\end{equation}
where $w_i$ is the label refinement coefficient. However, $w_i$ can be calculated from the JSD values as,
\begin{equation} \label{eq:weight}
  w_i  =\left\{
  \begin{array}{@{}ll@{}}
    1 - d_i, & \text{if} \hspace{0.2cm}  d_{i} \geq d_{\omega} \\
    1, & \hspace{0.05cm} \text{otherwise}
  \end{array}\right.
\end{equation}
where $d_{\omega}$ is the label-refinement threshold that adjusts $w_i$ based on the JSD of sample $\mathbf{x}_i$. Next, we follow the temperature sharpening \cite{li2020dividemix} step given that gives us $\hat{\mathbf{y}}_i$.

Similarly, we calculate pseudo-label by averaging the predictions of both networks \cite{li2020dividemix}, i.e.
\begin{equation} \label{eq:label_guessing}
     \bar{\mathbf{q}}_{b}  =\frac{1}{4}\sum_{m=1}^{2}\big(\mathbf{h}(\mathbf{f}(\hat{\mathbf{u}}^{weak}_{b,m};\theta^{(1)}); \phi^{(1)})+\\ \mathbf{h}(\mathbf{f}(\hat{\mathbf{u}}^{weak}_{b,m};\theta^{(2)}); \phi^{(1)})\big)
\end{equation}
and apply temperature sharpening on it to get $\mathbf{q}_b$.

We aggregate the labeled and unlabeled images with their ground-truth labels and pseudo-labels, respectively.  That is, $\hat{\mathcal{X}}=\{(\hat{\mathbf{x}}^{strong}_{i,m},\hat{\mathbf{y}}_i);i\in(1,...,N), m=(1,2)\}$, and $\hat{\mathcal{U}}=\{(\hat{\mathbf{u}}^{strong}_{i,m},\mathbf{q}_{i});i\in(1,...,N), m = (1,2) \}$ are the labeled and unlabeled sets. We use MixMatch \cite{berthelot2019mixmatch} to have
\begin{align}
\vspace{-6mm}
    \mathcal{W} = \shuffle\mathopen{}\big(\mathopen{}\concat(\hat{\mathcal{X}}, \hat{\mathcal{U}})\big), \\
    \hat{\mathcal{X}} = \big(\mathopen{}\mathrm{MixUp}(\hat{\mathcal{X}}_i, \mathcal{W}_i); i \in (1, \ldots, |\hat{\mathcal{X}}|)\big), \\
    \hat{\mathcal{U}} = \big(\mathopen{}\mathrm{MixUp}(\hat{\mathcal{U}}_i, \mathcal{W}_{i+|\hat{\mathcal{X}}|}); i \in (1, \ldots, |\hat{\mathcal{U}}|)\big). 
\vspace{-6mm}
\end{align}
MixUp \cite{zhang2018mixup} proposed a strategy for generating convex combination of two inputs: in this case, samples from labeled and unlabeled sets and their corresponding ground-truth labels and pseudo-labels.

\subsection{Loss Functions}
After applying MixMatch, the semi-supervised losses are calculated as follows \cite{berthelot2019mixmatch}, 
\begin{align}\label{eqn:l_combined}
    \mathcal{L}_\mathcal{X} &= \frac{1}{|\hat{\mathcal{X}}|} \sum_{\mathbf{x}, \mathbf{p} \in \hat{\mathcal{X}}} \xent(\mathbf{p}, \mathbf{h}(\mathbf{f}(\mathbf{y} \mid \mathbf{x}; \theta);\phi)) , \\
    \mathcal{L}_\mathcal{U} &= \frac{1}{|\hat{\mathcal{U}}|} \sum_{\mathbf{u}, \mathbf{q} \in \hat{\mathcal{U}}} \|\mathbf{q} - \mathbf{h}(\mathbf{f}(\mathbf{y} \mid \mathbf{u}; \theta);\phi)\|_2^2, \label{eqn:l_u} 
\end{align}
where $\xent(\mathbf{p}, \mathbf{q})$ is the cross-entropy between distributions $\mathbf{p}$ and $\mathbf{q}$ with $\mathbf{y}$ as the given label.  

Additionally, to prevent single-class assignment of all samples, we use  a regularization term based on a prior uniform distribution $(\pi_c = 1/C)$ to regularize the network's output across all samples in the mini-batch similar to Tanaka et al. \cite{tanaka2018joint} ,
\begin{equation}
    \mathcal{L}_{reg} = \sum_c {\pi_c log\big( \frac{\pi_c}{\frac{1}{|\hat{\mathcal{X}}+\hat{\mathcal{U}}|} \sum_{\mathbf{x} \in |\hat{\mathcal{X}}+\hat{\mathcal{U}}|} \mathbf{h}(\mathbf{f}(\mathbf{x}; \theta); \phi)} \big)}
\end{equation}
This gives us our semi-supervised loss function as shown in Figure \ref{fig:framework_sup},
\begin{equation} \label{eq:combined_loss}
    \mathcal{L}_{semi}=\mathcal{L}_\mathcal{X}+ \lambda_\mathcal{U} \mathcal{L}_\mathcal{U} + \lambda_r \mathcal{L}_\mathrm{reg}.    
\end{equation}
Here, $\lambda_\mathcal{U}$ and $\lambda_r$ are unsupervised loss coefficient and regularization coefficient, respectively.

\begin{table*}[htb]
\centering
\scalebox{0.9}{
\begin{tabular}{l|cc|cc|cc|cc} \toprule
Dataset & \multicolumn{4}{c}{CIFAR10} & \multicolumn{4}{|c}{CIFAR100} \\ \midrule
Noise Rate & \multicolumn{2}{c}{50\%} & \multicolumn{2}{|c}{80\%} & \multicolumn{2}{|c}{50\%} & \multicolumn{2}{|c}{80\%} \\ \midrule
Method & Best & Last & Best & Last & Best & Last & Best & Last \\ \midrule
\textsc{UniCon} w/o $\mathcal{L}_\mathcal{U}$ & 94.89 & 94.70  & 87.82 & 87.10 & 74.99 & 74.73 & 56.94 & 56.04\\
\textsc{UniCon} w/o $\mathcal{L}_{reg}$ & 95.38 & 95.11 & 93.59 & 93.26 & 76.48 & 75.87 & 61.75 & 60.90 \\ \midrule
\textsc{UniCon}  & \textbf{95.61} & \textbf{95.24} & \textbf{93.97} & \textbf{93.97} &  \textbf{77.63} & \textbf{76.91} & \textbf{63.98} & \textbf{63.13}\\ \bottomrule
\end{tabular}
}

\caption{Contribution of different loss functions on the performance of \textsc{UniCon}. While removing each loss term decreases the test accuracy, $\mathcal{L}_\mathcal{U}$ plays the most important role in obtaining SOTA performance. Test accuracies from the last epoch are also shown.}
\label{tab:loss_impact}
\end{table*}

\begin{table}[t]
\centering
\scalebox{0.9}{
\begin{tabular}{ll|cc|cc} \toprule
\multicolumn{2}{c|}{Loss Coef.} & \multicolumn{2}{c}{CIFAR10} & \multicolumn{2}{|c}{CIFAR100}  \\ \cline{3-6}
$\lambda_\mathcal{U}$ & $\lambda_\mathcal{C}$  &  Best & Last & Best & Last \\ \midrule
20 & 0.025 & 95.38 & 94.80 & 77.12 & 76.89 \\ 
30 & 0.025 & \textbf{95.61} & \textbf{95.24} & \textbf{77.63} & \textbf{76.91}\\
40 & 0.025 & 95.42 & 95.26 & 77.34 & 77.18 \\ 
20 & 0.050 &  95.49  & 94.83 & 77.46 & 76.95 \\ 
30 & 0.050 &  95.17  & 94.56 & 77.28 & 76.12  \\ 
40 & 0.050 &  95.35  & 94.79 & 77.15 & 76.44 \\ \bottomrule
\end{tabular}
}
\caption{Performance analysis of $\textsc{UniCon}$ with different loss coefficients (50\% symmetric noise). We observe that our proposed method is stable over different values of $\lambda_\mathcal{U}$ and $\lambda_\mathcal{C}$.}
\label{tab:loss_coeff}
\end{table}

\begin{figure*}[ht]
  \centering
  \begin{subfigure}{0.45\linewidth}
    \includegraphics[width=0.98\linewidth]{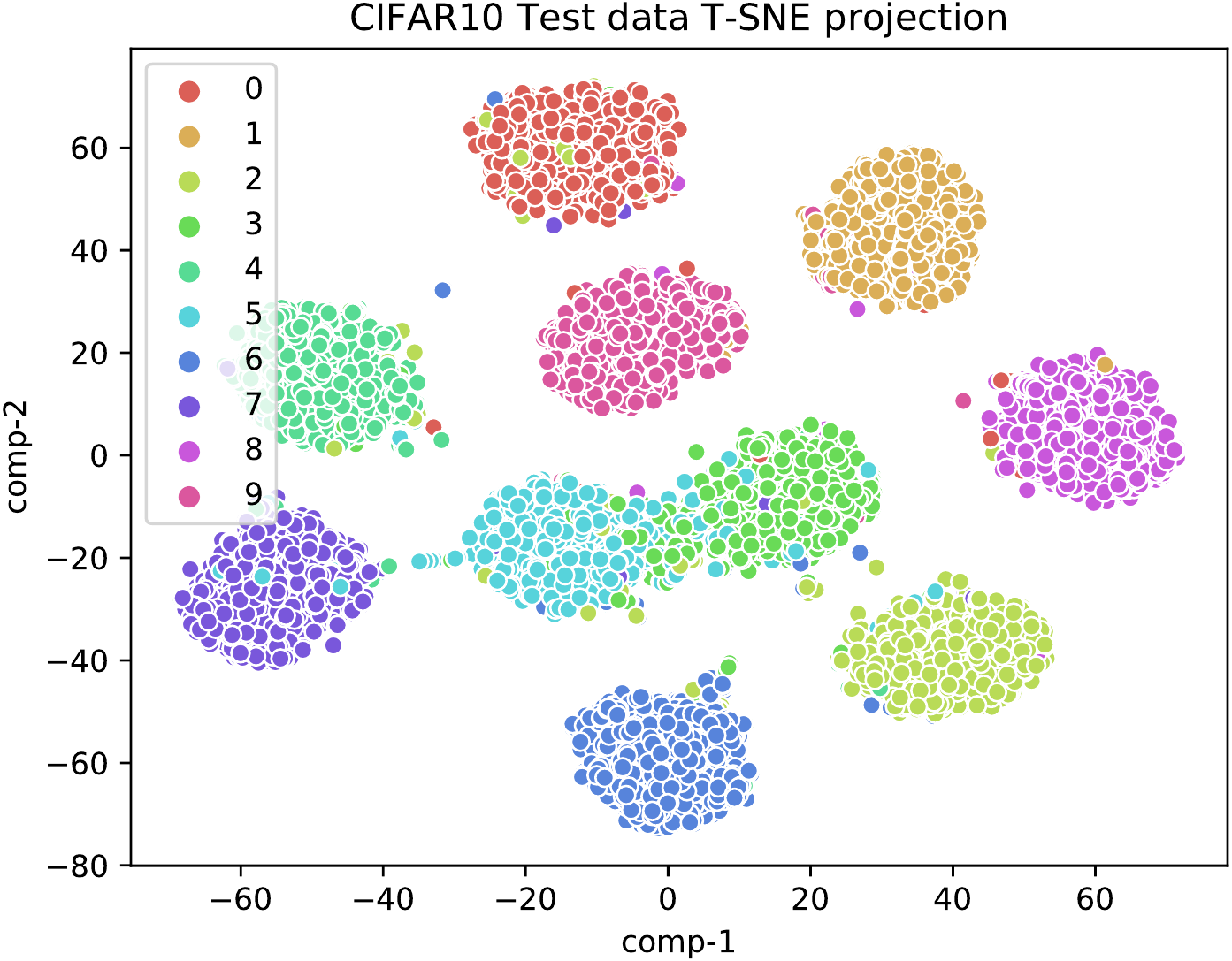}
    \caption{Symmetric 50\%}
    \label{fig:tsNE_50}
  \end{subfigure}
  \hfill
  \begin{subfigure}{0.45\linewidth}
    \includegraphics[width=0.98\linewidth]{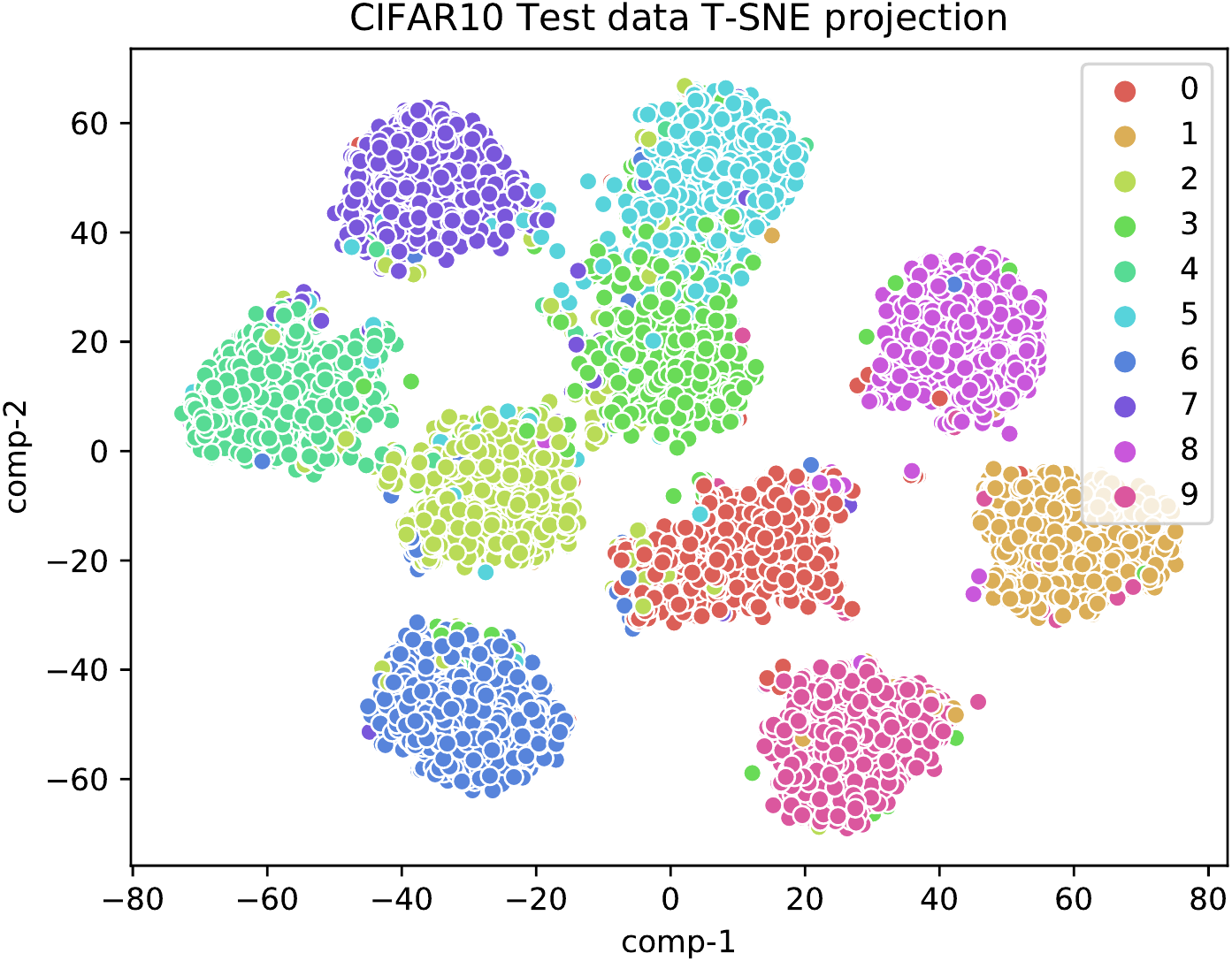}
    \caption{Symmetric 80\%}
    \label{fig:tSNE_80}
  \end{subfigure}

  \begin{subfigure}{0.45\linewidth}
  \vspace{5mm}
    \includegraphics[width=0.98\linewidth]{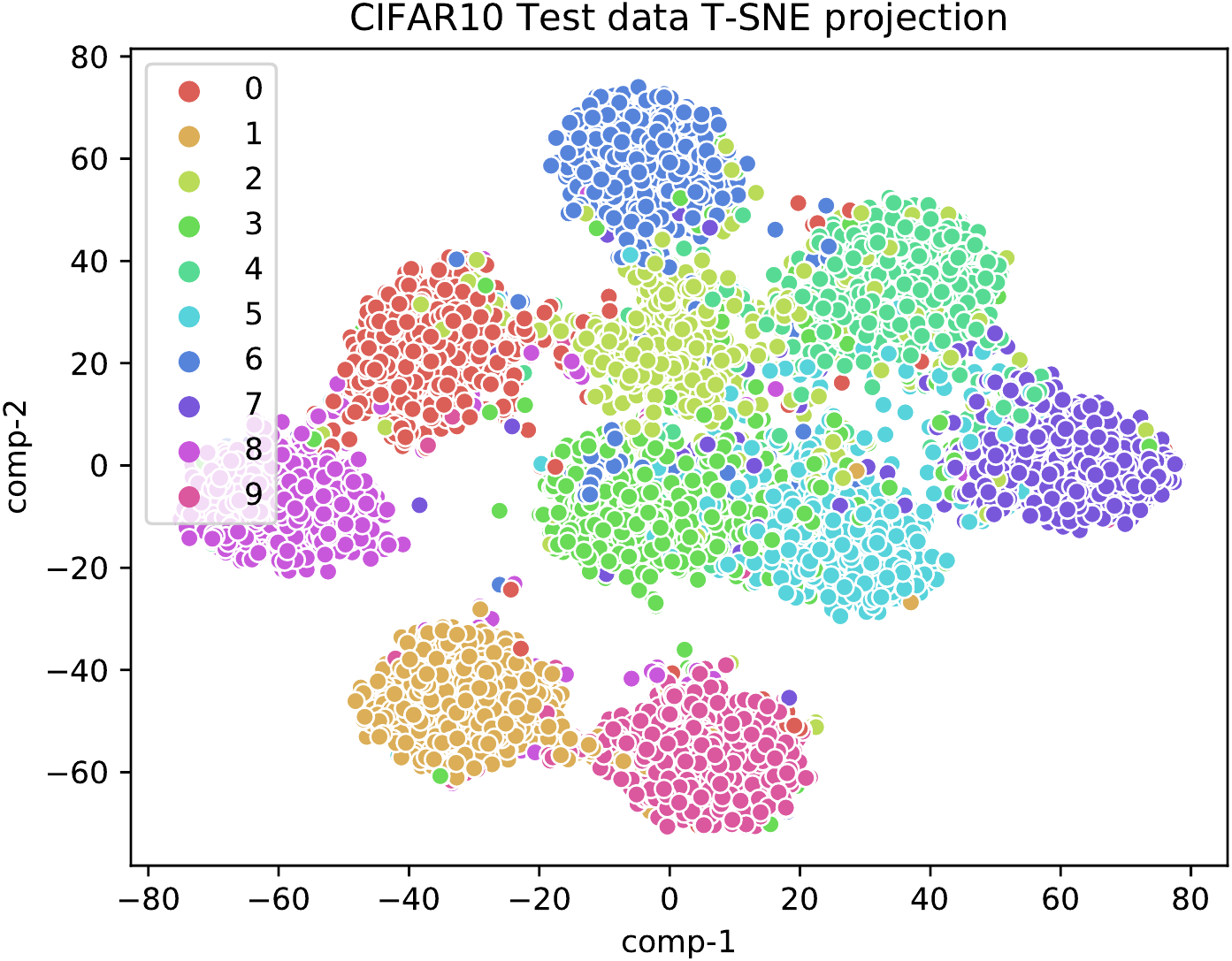}
    \caption{Symmetric 90\%}
    \label{fig:tSNE_90}
  \end{subfigure}
  \hfill
  \begin{subfigure}{0.45\linewidth}
  \vspace{5mm}
    \includegraphics[width=0.98\linewidth]{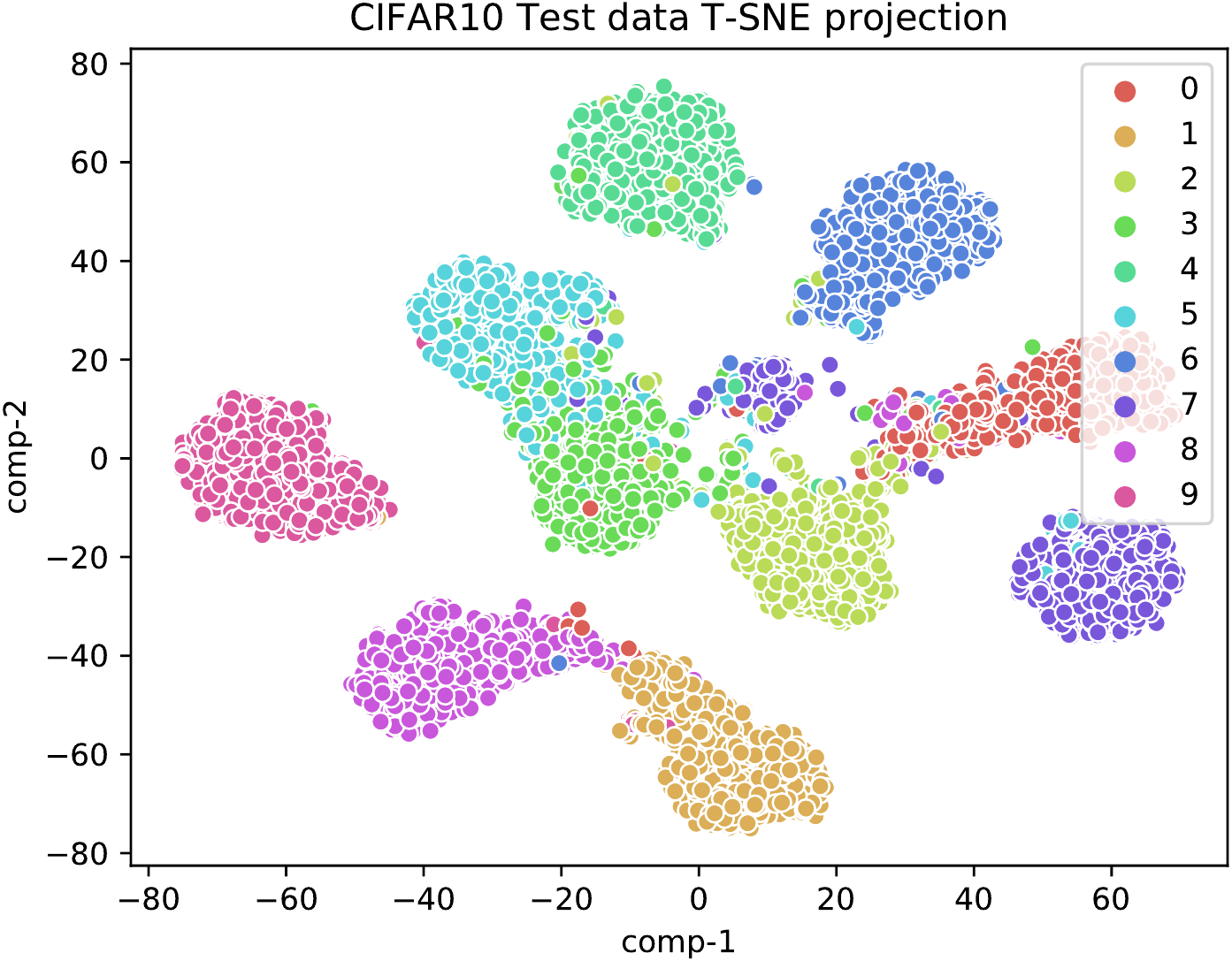}
    \caption{Asymmetric 40\%}
    \label{fig:tSNE_40}
  \end{subfigure}
  \caption{T-SNE visualizations of network features of test images. The graphs show class distribution after training the network for 300 epochs on CIAFAR10 dataset with different noise types: (a) 50\% symmetric, (b) 80\% symmetric, (c) 90\% symmetric, (d) 40\% asymmetric. Even under extreme label-noise, \textsc{UniCon} effectively learns the true class distributions.}
  \label{fig:tSNE}
\end{figure*}

\begin{figure*}[ht]
  \centering
  \begin{subfigure}{0.45\linewidth}
  \vspace{5mm}
    \includegraphics[width=0.98\linewidth]{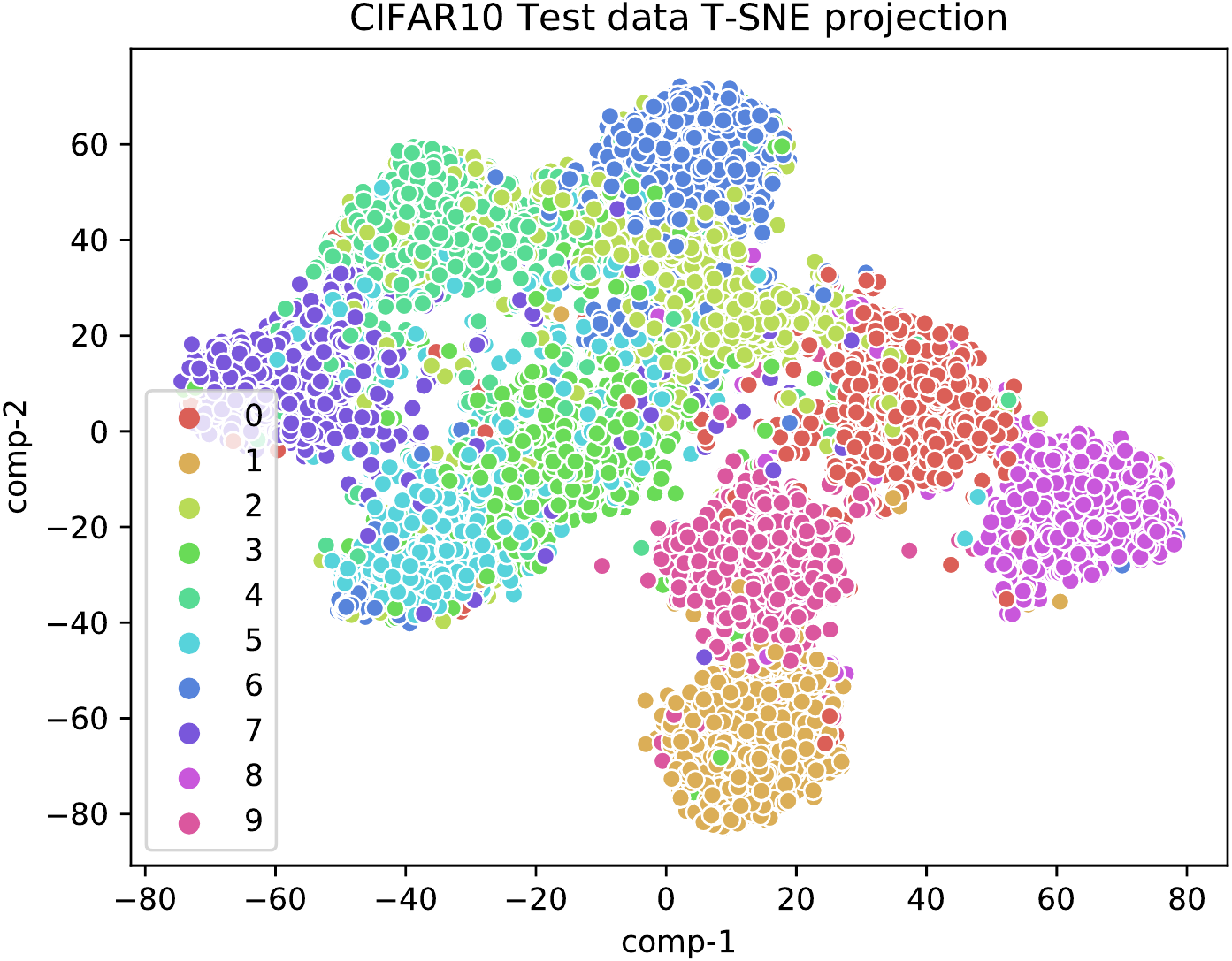}
    \caption{Class distribution learned by \textsc{UniCon}}
    \label{fig:tsNE_uni}
  \end{subfigure}
  \hfill
  \begin{subfigure}{0.45\linewidth}
  \vspace{5mm}
    \includegraphics[width=0.98\linewidth]{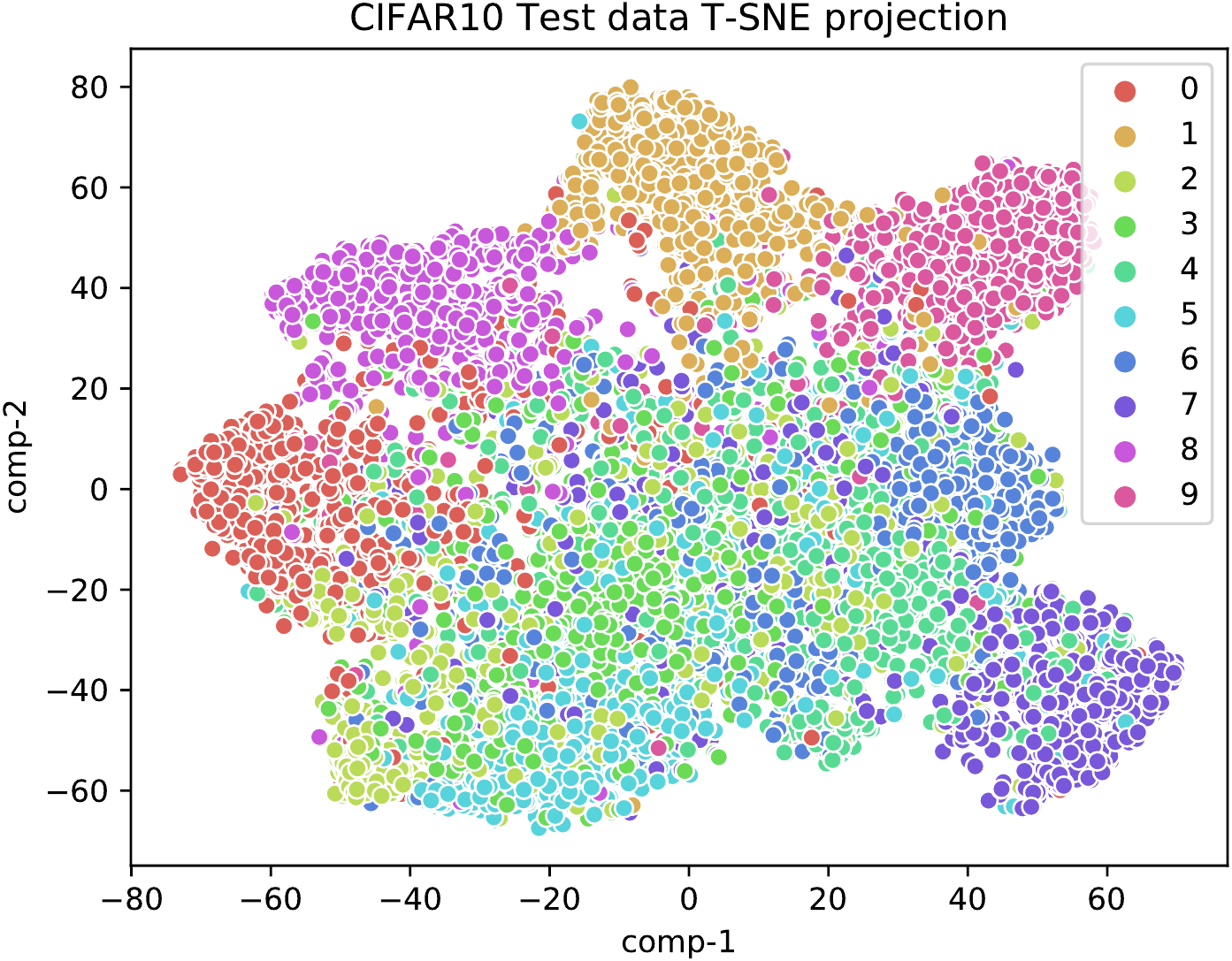}
    \caption{Class distribution learned by DMix~\cite{li2020dividemix}}
    \label{fig:tSNE_div}
  \end{subfigure}
  \caption{Class distribution learned by (a) the proposed \textsc{UniCon}  and (b) DMix~\cite{li2020dividemix} on CIFAR10 dataset with 95\% symmetric noise. \textsc{UniCon} shows better class separation even when only 5\% samples have correct labels.}
  \label{fig:tSNE_95}
\end{figure*}

\begin{figure}
    \centering
    \includegraphics[width=0.98\linewidth]{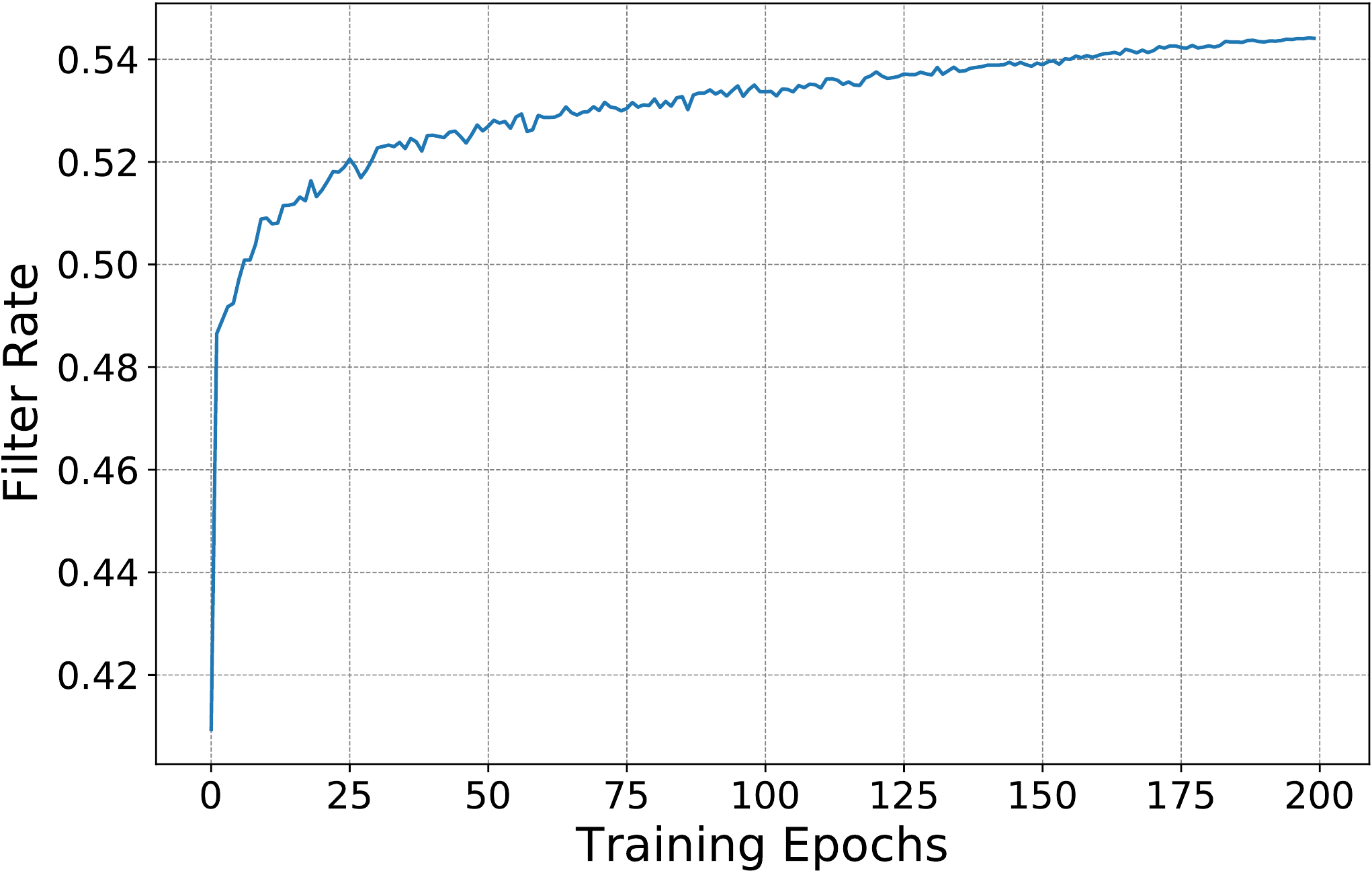}
    \caption{Our designed filtering rate $R$ adjusts itself based on the network predictions without manual tuning at each training iteration  \cite{yao2021jo}. As training progresses and the model gets confident about most of its predictions, \textsc{UniCon} selects more clean samples with better precision. For this graph we used CIFAR10 dataset with 50\% symmetric noise.}
    \label{fig:filter_rate}
\end{figure}
\begin{algorithm*}[htb]
	\DontPrintSemicolon
	\small
	\textbf{Input}: network-1 parameters $\Theta^{(1)} = (\theta^{(1)}, \phi^{(1)}, \psi^{(1)})$  and network-2 parameters $\Theta^{(2)}=(\theta^{(2)}, \phi^{(2)}, \psi^{(2)})$, training set $\mathbb{D} = (\mathcal{X},\mathcal{Y})$, number of samples $N$, number of classes $C$,
	sharpening temperature $T$, unsupervised loss coefficient $\lambda_\mathcal{U}$, contrastive loss coefficient $\lambda_\mathcal{C}$, and regularization coefficient $\lambda_\mathrm{r}$ .  
	
	\For {$k=1$ \KwTo $2$}
	{
        $\mathbb{D}_{clean}, \mathbb{D}_{noisy}, \boldsymbol{d} = \textit{Uniform\textendash Selection}\big(\mathbb{D}, (\theta^{(1)}, \phi^{(1)}), (\theta^{(2)}, \phi^{(2)}), N, C \big)$ (see Alg. 1 of main paper) \tcp*{Separation of clean and noisy set} 
        $\mathbb{W} = \textit{{Weight\textendash Estimation}}\big(\boldsymbol{d}\big) $ (see eq. \ref{eq:weight}) \tcp*{Weights for label-refinement} 
		\For {$\mathrm{iter}=1$ \KwTo $\mathrm{num\_iters}$}
		{
			From $(\mathbb{D}_{clean}, \mathbb{W})$, draw a mini-batch $\{(\mathbf{x}_b,\mathbf{y}_b, w_b);b\in(1,...,B)\}$ \tcp*{Draw labeled data for SSL}
			From $\mathbb{D}_{noisy}$, draw a mini-batch $\{\mathbf{u}_b; b\in(1,...,B)\}$ \tcp*{Draw unlabeled data for SSL}
			\For{$b=1$ \KwTo $B$}
			{   
                \For{$m=1$ \KwTo $2$}     
                {
                    $\hat{\mathbf{x}}^{weak}_{b,m}=\textit{Weak\textendash Augment}(\mathbf{x}_b)$ \tcp*{First weakly-augmented copy} $\hat{\mathbf{u}}^{weak}_{b,m}=\textit{Weak\textendash Augment}(\mathbf{u}_b)$  \tcp*{Second weakly-augmented copy}
        			$\hat{\mathbf{x}}^{strong}_{b,m}=\textit{Strong\textendash Augment}(\mathbf{x}_b)$ \tcp*{First strongly-augmented copy}
    				$\hat{\mathbf{u}}^{strong}_{b,m}=\textit{Strong\textendash Augment}(\mathbf{u}_b)$  \tcp*{Second-strongly augmented copy}
                }
				Get ${\mathbf{p}}_{b}$ using Eq. \ref{eq:label_re} \tcp*{Model Prediction}
				$\bar{\mathbf{y}}_b=w_b \mathbf{y}_b+(1-w_b){\mathbf{p}}_{b}$ \tcp*{Label-refinement} 
				$\hat{\mathbf{y}}_b=\mathrm{Sharpen}(\bar{\mathbf{y}}_b,T)$ 	\tcp*{Temperature sharpening} 
				Get $\bar{\mathbf{q}}_{b}$ using Eq. \ref{eq:label_guessing}
				 \tcp*{Pseudo-label} 
				${\mathbf{q}}_{b}=\mathrm{Sharpen}(\bar{\mathbf{q}}_{b},T)$ \tcp*{Temperature sharpening} 
			}
			$\hat{\mathcal{X}}=\{(\hat{\mathbf{x}}^{strong}_{b,m},\hat{\mathbf{y}}_b);b\in(1,...,B)\}$ \tcp*{labeled Set} 
			$\hat{\mathcal{U}}=\{(\hat{\mathbf{u}}^{strong}_{b,m},\mathbf{q}_{b});b\in(1,...,B)\}$ \tcp*{Unlabeled Set} 
	        $\mathcal{L}_\mathcal{X},\mathcal{L}_\mathcal{U}=\mathrm{MixMatch}(\hat{\mathcal{X}},\hat{\mathcal{U}})$ \tcp*{Apply MixMatch} 
	        
	        Calculate $\mathcal{L}_\mathcal{C}$ using eq. \ref{eq:con_loss}  \tcp*{Contrastive Loss}
			$\mathcal{L}_{tot}=\mathcal{L}_\mathcal{X}+ \lambda_\mathcal{U} \mathcal{L}_\mathcal{U} + \lambda_\mathcal{C} \mathcal{L}_\mathcal{C} + \lambda_r \mathcal{L}_\mathrm{reg}$ \tcp*{Total loss} 
			$\Theta^{(k)}=\mathrm{SGD}(\mathcal{L},\Theta^{(k)})$ \tcp*{Update the Parameters}
		}
	}
	$\textbf{Return:}$ Updated $\Theta^{(1)}$, $\Theta^{(2)}$.
	\caption{\small One epoch of SSL Training}
	\label{alg:SSL_training}
	
\end{algorithm*}

We consider another loss function, contrastive loss, which is used only for the data points in $\mathbb{D}_{noisy}$. 
Let the projection head output corresponding to  $\hat{\mathbf{u}}^{strong}_{i,1}$ and $\hat{\mathbf{u}}^{strong}_{i,2}$ be $\mathbf{z}_{i}$ and $\mathbf{z}_{j}$, respectively. The contrastive loss function \cite{chen2020simple,khosla2020supervised} can be defined as
\begin{align}
        \ell_{i,j}  & = -\log \frac{\exp(\mathrm{sim}( \mathbf{z}_{i}, \mathbf{z}_{j})/\kappa)}{\sum_{b=1}^{2B} \mathbbm{1}_{b \neq i}\exp(\mathrm{sim}(\mathbf{z}_i, \mathbf{z}_b)/\kappa)}~, \\
        & \mathcal{L}_\mathcal{C}= \frac{1}{2B}\sum_{b=1}^{2B}[\ell_{2b-1,2b} + \ell_{2b,2b-1}] \label{eq:con_loss},
\end{align}
where $\mathbbm{1}_{b \neq i}$ is an indicator function that gives a 1 iff $b \neq i$, $\kappa$ is a temperature constant, $B$ is the number of samples in mini-batch, and $\mathrm{sim}(\mathbf{z}_{i}, \mathbf{z}_{j})$ can be expressed as the cosine similarity between $\mathbf{z}_{i}$ and $\mathbf{z}_{j}$. 

For each mini-batch, there are total 2B augmented samples, since we are creating a pair of augmented samples out of a single sample.  Let us consider $i$ and $j$ as a positive pair, then the rest of the data points ($2B-2$) are treated as negative examples. We can compute the final contrastive loss $\mathcal{L}_\mathcal{C}$ across all the positive pairs, both ($i,j$) and ($j,i$) in a single mini-batch. The formulation of $ \ell_{i,j} $ does not require any labels (ground-truth or pseudo-labels). Since contrastive loss does not require labels, it mitigates the negative impact of noisy label memorization. 

\begin{table*}[t]
\centering
\scalebox{1}{
\begin{tabular}{|l|c|c|c|c|}
\toprule
\textbf{Hyper Parameters} & \textbf{CIFAR10/100} & \textbf{Tiny-ImageNet200} & \textbf{Clothing1M} & \textbf{WebVision} \\
\midrule
Optimizer & SGD & SGD & SGD & SGD  \\
\midrule
Initial Learning Rate & $0.02$ & $0.01$ & $0.002$ & $0.01$ \\
Momentum     & 0.9 & 0.9 & 0.9 & 0.9 \\
Weight Decay & $5e^{-4}$ & $5e^{-4}$ & $1e^{-3}$ & $1e^{-3}$ \\
Mini-batch Size & 64 & 32 & 32 & 32 \\
Total Epochs & 300/350 & 350 & 8 & 100 \\
$T$ & 0.5 & 0.5 & 0.5 & 0.5 \\
$\lambda_\mathcal{C}$ & 0.025 & 0.025 & 0.025 & 0.025 \\
$\lambda_\mathcal{U}$ & 30 & 30 & 30 & 30 \\
$\lambda_r$ & 1 & 1 & 1 & 1\\
$\kappa$ & 0.05 & 0.05 & 0.05 & 0.05 \\
$d_{\omega}$ & 0.5 & 0.5 & 0.5 & 0.5 \\
MixUp, $\alpha$ & 4 & 2 & 0.5 & 0.5 \\
\bottomrule
\end{tabular}}
\caption{Hyperparameter Settings for \textsc{UniCon}. Most of the parameters are the same across different datasets. This shows the general applicability of the proposed \textsc{UniCon} method.}
\label{tab:hyperparameters}
\end{table*}

Finally, we accumulate all losses to get the total loss, 
\begin{equation}
    \mathcal{L}_{tot}=\mathcal{L}_{semi} + \lambda_\mathcal{C} \mathcal{L}_\mathcal{C},    
\end{equation}
where $\lambda_\mathcal{C}$ is the contrastive loss coeffiecient. The summary of these steps is provided in Algorithm.~\ref{alg:SSL_training}.

\section{Ablation Studies} \label{sec:ablation_studies}
In this section, we analyze the performance of \textsc{UniCon} under different scenarios.

\subsection{Impact of Different Losses}
We observe the contribution of each loss function on the performance of \textsc{UniCon}. It can be observed from Table \ref{tab:loss_impact} that each loss term helps in improving the performance while $\mathcal{L}_\mathcal{U}$ has the highest impact on performance. Training without $\mathcal{L}_\mathcal{U}$ indicates that we discard the selected noisy samples completely. The drop in accuracy shows the significance of pseudo-label based feature learning. Improving the quality of these pseudo-labels is one of the primary contributions of \textsc{UniCon}.    


\subsection{Loss Coefficients}
In Table \ref{tab:loss_coeff}, we show the effect of different loss coefficients. We observe that the performance of \textsc{UniCon} is relatively stable over a large range of coefficient values. We select a value of 30 and 0.025 for $\lambda_\mathcal{U}$ and $\lambda_\mathcal{C}$ respectively since this set of values result in optimal performance on both CIFAR10 and CIFAR100 datasets. We apply the same loss coefficient value for all datasets irrespective of the class number, number of samples, noise type, noise rate etc.

\subsection{T-SNE Visualization}
A t-SNE visualization\cite{van2008visualizing} for features of test images is presented in Figure \ref{fig:tSNE}. The features are obtained from models trained under different label noise settings. We observe that class separation gets better as the noise level decreases. We further notice that \textsc{UniCon} obtains the best separation of test images at symmetric $50\%$ noise. However, when the noise rate increases it becomes more challenging to learn the class distribution as shown in Figure. \ref{fig:tSNE_80} and \ref{fig:tSNE_90}. In addition, we compare the performance of our method with DMix~\cite{li2020dividemix} in the presence of 95\% label noise in Figure~\ref{fig:tSNE_95}. It is a difficult task to separate clean samples from noisy samples under such high noise rate. Interestingly, we observe that our simple approach effectively learns better class distribution in comparison to DMix~\cite{li2020dividemix}. We attribute this to the high precision of our uniform clean sample selection strategy.

\subsection{Memorization of Noisy Labels}
In case of standard training, the network memorizes the noisy labels leading to poor generalization performance. However, our proposed method \textsc{UniCon} demonstrates resistance to memorization of label noise. We show this phenomena in Figure \ref{fig:memorization}. We observe that with standard training the accuracy improves consistently over different epochs suggesting the memorization of label noise. In sharp contrast to this, the training accuracy of \textsc{UniCon} saturates very quickly indicating that the network is resisting the memorization of noisy labels at later stage of training. For instance, an ideal scenario for 80\% symmetric noise would be if the training accuracy is $\sim$20\%, i.e. the percentage of clean samples. Furthermore, we notice that our training accuracy deteriorates as we increase the rate of label noise in the training data.  This further validates our claim that \textsc{UniCon} is effective in combating the memorization of label noise.  

\subsection{Filter Rate}
In Figure \ref{fig:filter_rate}, we show that the filter rate steadily increases as the network generates more confident predictions (shown for 50\% noise rate). At each epoch of training, the filter rate, $R$ is selected based on network predictions. This design decision omits the requirement of manually tuning the selection parameter (filter rate) at each training epoch \cite{yao2021jo}.  For our experiments, we set $d_{\mu}$, and $\tau$ to 0.7, and 5 respectively.  

\section{Baseline Methods}
For CIFAR10 and CIFAR100, we compare UniCon with the following state-of-the art methods: LDMI \cite{xu2019ldmi}, M-Up \cite{zhang2018mixup}, PCIL \cite{yi2019probabilistic}, ELR \cite{liu2020earlylearning}, DMix \cite{li2020dividemix}, MOIT \cite{ortego2021multiobjective}. Methods like ELR \cite{liu2020earlylearning} focus on the importance of the early learning regularization in preventing the memorization; MOIT \cite{ortego2021multiobjective} porposes a multi-objective framework to deal with the noisy labels. 
For Clothing1M, we consider Joint-Optim \cite{tanaka2018joint}, MetaCleaner \cite{zhang2019metacleaner} along with ELR \cite{liu2020earlylearning} and DMix  \cite{li2020dividemix} . Furthermore, D2L \cite{ma2018dimensionality}, MentrorNet \cite{jiang2018mentornet} 
Co-Teaching \cite{han2018co}, Iterative-CV \cite{wang2018iterative} are among the methods we consider for WebVision. For TinyImageNet, we compare our method with Decoupling~\cite{malach2017decoupling},  MentorNet~\cite{jiang2018mentornet}, Co-teaching+~\cite{yu2019does}, M-correction~\cite{arazo2019unsupervised}, NCT \cite{sarfraz2020noisy} etc. 

\section{Training Details} \label{sec:training_details}
\subsection{Hyper-parameter Settings}
We describe the hyperparameter settings in Table \ref{tab:hyperparameters}. Note that most of these hyperparameters are the same across all datasets. 

\subsection{WebVision and Clothing1M}
For Clothing1M dataset, first, we resize the image to $256 \times 256$ and then apply random crop to those images to obtain a $224 \times 224$ image. On the other hand, each image of WebVision is resized to $320 \times 320$ and a random crop of size $299 \times 299$ is applied. For WebVision, we consider only 50 classes for training and validation. Similarly, only 50 classes are considered for ILSVRC12 validation set. The percentage of noisy labels in WebVision are estimated to be around $20\%$. It has been shown that our method obtains slightly lower \emph{Top-1} accuracy than state-of-the-art.  In some scenarios (low noise level), the experimental results indicate that \textsc{UniCon} underperforms compared to the state-of-the-art. Relatively low performance on WebVision dataset can be attributed to the presence of low label noise. 

\end{document}